\documentclass[conference]{IEEEtran}
\IEEEoverridecommandlockouts

\newcommand{\ourmethod}{\emph{Grasp-MPC}}
\usepackage{algorithm}
\usepackage[noend]{algpseudocode}
\usepackage{float}
\usepackage{bbm}
\usepackage{graphicx}
\usepackage{subcaption}
\usepackage{pgfplots}
\pgfplotsset{compat=1.18}
\usepgfplotslibrary{groupplots}

\usepackage{hyperref}       %
\usepackage{url}            %
\usepackage{booktabs}       %
\usepackage{amsfonts}       %
\usepackage{amsmath,amssymb}
\usepackage{nicefrac}       %
\usepackage{amsmath,amsfonts,bm}
\usepackage{wrapfig}
\usepackage{mathtools}
\usepackage[table]{xcolor}
\usepackage{multirow}
\usepackage{listings}
\usepackage{cuted}
\usepackage{capt-of}

\newcommand{\Skip}[1]{}

\newcommand{\RNum}[1]{\uppercase\expandafter{\romannumeral #1\relax}}

\definecolor{codegreen}{rgb}{0,0.6,0}
\definecolor{codegray}{rgb}{0.5,0.5,0.5}
\definecolor{codepurple}{rgb}{0.58,0,0.82}
\definecolor{backcolour}{rgb}{0.95,0.95,0.92}

\lstdefinestyle{mystyle}{
    backgroundcolor=\color{backcolour},   
    commentstyle=\color{codegreen},
    keywordstyle=\color{magenta},
    numberstyle=\tiny\color{codegray},
    stringstyle=\color{codepurple},
    basicstyle=\ttfamily\footnotesize,
    breakatwhitespace=false,         
    breaklines=true,                 
    captionpos=b,                    
    keepspaces=true,                 
    numbers=left,                    
    numbersep=5pt,                  
    showspaces=false,                
    showstringspaces=false,
    showtabs=false,                  
    tabsize=2
}

\def\eqref#1{equation~\ref{#1}}

\def\1{\bm{1}}

\DeclareMathAlphabet{\mathsfit}{\encodingdefault}{\sfdefault}{m}{sl}
\SetMathAlphabet{\mathsfit}{bold}{\encodingdefault}{\sfdefault}{bx}{n}

\DeclareMathOperator*{\argmin}{arg\,min}

\title{\LARGE \bf
Grasp-MPC: Closed-Loop Visual Grasping via \\ Value-Guided Model Predictive Control

\author{Jun Yamada$^{1,2}$ \quad
Adithyavairavan Murali$^2$ \quad
Ajay Mandlekar$^2$ \\
Clemens Eppner$^2$ \quad
Ingmar Posner$^1$ \quad
Balakumar Sundaralingam$^2$
}

\thanks{Work done during internship at NVIDIA.}
\thanks{$^1$Applied AI Lab, Oxford Robotics Institute, University of Oxford}
\thanks{$^2$NVIDIA, USA}

\thanks{Correspondence to: {\tt\small jyamada@robots.ox.ac.uk}}%

}

\begin{document}

\maketitle
\thispagestyle{empty}
\pagestyle{empty}

\begin{strip}
\vspace{-2.0cm}
\centering
\includegraphics[width=\linewidth]{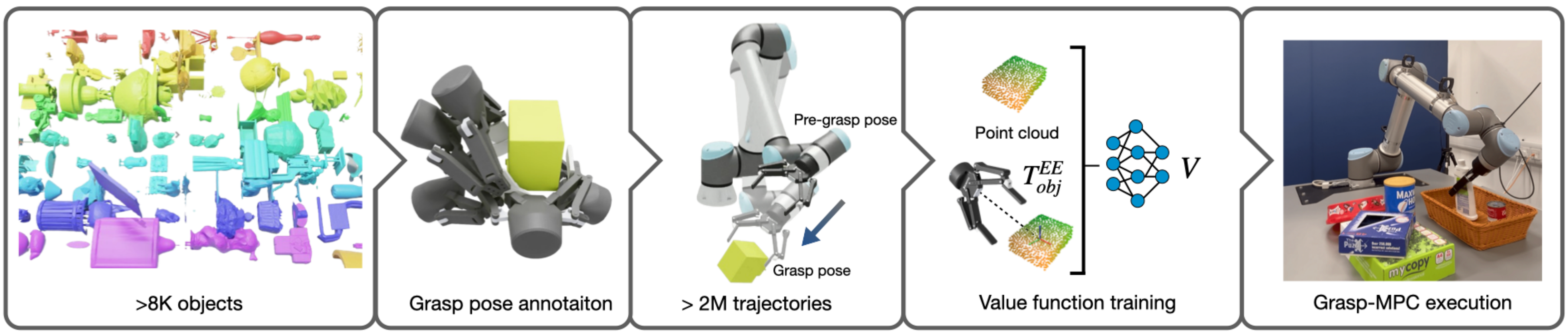}
\captionof{figure}{
\ourmethod~overview. A large-scale synthetic grasp trajectory dataset is generated in simulation using a motion planner, collecting only trajectories between pre-grasp and ground-truth grasp poses across $8$K Objaverse objects. A value function is trained using a sparse cost label given the target object’s point cloud and the end-effector pose. The learned value function is used in an MPC framework, enabling robust and safe grasping of novel objects in cluttered environments.}
\label{fig:teaser}
\vspace{-0.5cm}
\end{strip}

\begin{abstract}
Grasping of diverse objects in unstructured environments remains a significant challenge.
Open-loop grasping methods, effective in controlled settings, struggle in cluttered environments. 
Grasp prediction errors and object pose changes during grasping are the main causes of failure.
In contrast, %
closed-loop methods address %
these challenges %
in simplified settings (e.g., single object on a table) on a limited set of objects, with no path to generalization.
We propose \ourmethod, a closed-loop 6-DoF vision-based grasping policy designed for robust and reactive grasping %
of novel objects in cluttered environments.
\ourmethod~incorporates a value function, trained on visual observations from a large-scale synthetic dataset of 2 million grasp trajectories that include successful and failed attempts. We deploy this learned value function in an MPC framework in combination with other cost terms that encourage collision avoidance and smooth execution.
We evaluate \ourmethod~on FetchBench and real-world settings across diverse environments. 
\ourmethod~improves grasp success rates by up to 32.6\% in simulation and 33.3\% in real-world noisy conditions, outperforming open-loop, diffusion policy, transformer policy, and IQL approaches.
Videos and more at \url{http://grasp-mpc.github.io}.
\end{abstract}

\section{Introduction}

Grasping is a foundational capability in robotics, serving as a prerequisite for physical interaction and subsequent complex manipulation tasks.
However, despite decades of research, grasping remains unsolved, particularly when dealing with novel objects in cluttered scenes.
State-of-the-art grasping methods can broadly be categorized into open-loop and closed-loop approaches, but both fall short of meeting the requirements of robust grasping in unstructured environments.
Open-loop methods, which rely on grasp pose prediction models~\cite{yuanm2t2, mousavian20196, sundermeyer2021contact, carvalho2024graspdiffusionnetworklearning}, use motion planning to reach predicted grasp poses.
Although demonstrating notable performance in grasping novel objects, these approaches are inherently unable to incorporate real-time feedback to adjust their goals, making them sensitive to grasp pose prediction errors and changes in object pose.
On the other hand, closed-loop policies, including those based on reinforcement learning (RL) and imitation learning (IL), address some of these shortcomings by incorporating feedback during execution.
However, prior approaches~\cite{kalashnikov2018scalable, wang2022goal, singh2024dextrah, lum2024dextrahg} are often limited to simplified settings, such as tabletop environments, and exhibit poor generalization to novel objects, primarily due to the absence of large-scale grasp trajectory datasets for diverse objects.
More importantly, safety considerations, such as collision avoidance in cluttered scenes, remain largely unaddressed during development.

To address these, we introduce \ourmethod, a framework that combines the strengths of open-loop and closed-loop methods for 6-DoF grasping in cluttered, novel-object settings.
\ourmethod~unifies model- and data-driven approaches by leveraging model predictive control (MPC) with a value function learned from data as a task cost function.
\ourmethod~uses a grasp prediction model and motion planning, similar to open-loop methods, to reach (noisy) pre-grasp poses.
It addresses prediction errors and object pose changes by employing MPC for closed-loop execution, enabling real-time feedback while enforcing constraints such as collision avoidance and minimum jerk.

Furthermore, \ourmethod~addresses a key challenge in applying MPC to grasping: designing a cost function that meaningfully captures grasp success. 
Traditional geometric cost functions based on distances to predicted grasp poses are sensitive to prediction errors and fail to exploit MPC’s full closed-loop capability. 
To overcome this, \ourmethod~leverages a vision-based value function, trained on large-scale synthetic grasping trajectories with Objaverse objects~\cite{deitke2024objaverse} using visual observations and sparse success labels (Figure~\ref{fig:teaser}). 
This value function predicts the likelihood of grasp success and serves as a task cost in MPC, guiding the robot to explore the state space toward successful grasps.

To this end, this work makes the following contributions:

\begin{enumerate}
    \item \ourmethod, a safe closed-loop visual grasping policy for novel objects in cluttered environments.
    \item \ourmethod~unifies model-based control and data-driven approach by integrating a learned grasp value function into an MPC framework, enabling reactive, constraint-aware grasp execution in dynamic and cluttered environments.    
    \item A large-scale synthetic grasp trajectory dataset comprising over $2$M trajectories, $115$M states, and $8515$ unique Objaverse objects, supporting scalable learning of a generalizable value function.
    \item Extensive real-world and simulated evaluations show that \ourmethod~significantly outperforms SOTA open-loop and closed-loop methods.
\end{enumerate}
\section{Related Works}
\textbf{Grasping.}
A significant body of research has focused on learning grasp prediction models~\cite{yuanm2t2, sundermeyer2021contact, mousavian20196, barad2024graspldm, fang2020graspnet}, typically paired with motion planning for execution.
These methods often rely on synthetic datasets with grasp annotations~\cite{eppner2021acronym}, enabling scalable training in simulation.
While effective for novel objects, open-loop approaches suffer from prediction errors and lack feedback integration, limiting real-world robustness.
In contrast, closed-loop policies learned via RL~\cite{kalashnikov2018scalable, wang2022goal, singh2024dextrah, lum2024dextrahg} and IL~\cite{Song2019GraspingIT, han2024fetchbench} address these issues, but often face challenges with sample inefficiency, limited generalization, and perform only in a clean tabletop.

Prior work~\cite{yamada2021motion, yamada2023efficient, han2024fetchbench} combines motion planning with learned policies for obstacle-aware manipulation.
FetchBench~\cite{han2024fetchbench} uses a Transformer-based IL policy~\cite{dalal2023imitatingtaskmotionplanning} with motion planning for grasping in clutter, but performance is limited by dataset size and diversity.
In contrast, we generate a large-scale grasping trajectory dataset, $100\times$ larger than~\cite{han2024fetchbench}, enabling better generalization.

\textbf{Model Predictive Control for Robotic Manipulation.}
Model Predictive Control (MPC)\cite{Abbeel2010AutonomousHA, williams2016, di2018} is a powerful framework for robotic control, enabling closed-loop optimization using real-time feedback.
However, applying MPC to grasping is challenging, as it requires a cost function that captures the nuances of grasp success~\cite{chen2022neural}.
Even with a target grasp pose from a prediction model, MPC remains vulnerable to errors, much like open-loop methods.
Recent work integrates learning with MPC to tackle challenges by learning dynamics~\cite{Lenz2015DeepMPCLD, finn2017, wahlstrom2015pixels, watter2015embed, ebert2018visual, pmlr-v97-hafner19a, hansen2022temporal}, cost functions~\cite{jawale2024dynamicnonprehensileobjecttransport,zhong2013, hansen2022temporal, lowrey2018plan, chen2022neural}, and sampling distributions~\cite{sacks2023learning, yamada2024dcubedlatentdiffusiontrajectory}.
Chen et al.~\cite{chen2022neural} propose a distance-based value function using a predicted grasp pose, but this overlooks key factors for grasp success, leading to suboptimal results.
CV-MPC~\cite{jawale2024dynamicnonprehensileobjecttransport} learns value ensembles from few demonstrations but rely on low-dimensional states, limiting generalization.
In contrast, \ourmethod~learns a value function from large-scale synthetic trajectories using point cloud observations and sparse success/failure labels, addressing these limitations.

\textbf{Offline RL}
Offline RL~\cite{peng2019advantage, kostrikov2021offline, kumar2020conservative} enables policy learning using offline datasets without environment interaction, leveraging both successful and failure trajectories to improve policies.
However, policy extraction often represents a bottleneck in the learning process rather than value function estimation, as discussed in prior work~\cite{park2024value}.
On the other hand, \ourmethod~does not need to extract the policy from the learned value function because \ourmethod~uses MPC as a policy that can explore and exploit states to minimize the learnt cost represented by the value function.

\section{Preliminary}
\textbf{Problem Formulation.}
We formulate grasping as a Partially Observable Markov Decision Process (POMDP).
A trajectory is defined as $\tau = (\mathbf{x}_{t}, \mathbf{a}_{t}, c_{t}, \mathbf{x}_{t+1}, \mathbf{a}_{t+1}, c_{t+1}, \dots)$, where $\mathbf{x} \in \mathcal{X}$ are observations, $\mathbf{a} \in \mathcal{A}$ actions, and $c \in \mathcal{C}$ costs.
The offline dataset contains $N$ trajectories $\{\tau^{i}\}^{N}_{i=1}$, comprising both successes and failures.
The objective is to minimize the discounted cumulative cost $J(\tau)=\sum{t'=t}^{\infty} \gamma^{t-t'} c(\mathbf{x}_{t}, \mathbf{a}_{t})$, with discount factor $\gamma$.

\textbf{Dynamics Model for Model Predictive Control.}
MPC samples action sequences and plans future states to select the next action that minimizes cost in real time, given a dynamics model.
In this work, MPC is optimizing for joint accelerations with Euler integration as the dynamics model. We assume that the real robot can track the generated joint position, velocity and acceleration target accurately with an inverse dynamics controller, similar to prior work~\cite{bhardwaj2022storm}.
The environment is not explicitly modelled; therefore, its dynamics, including those of objects, are unknown a priori.
\section{Approach}
\begin{figure*}[t]
    \vspace{-0.5cm}
    \centering
    \includegraphics[width=0.85\linewidth]{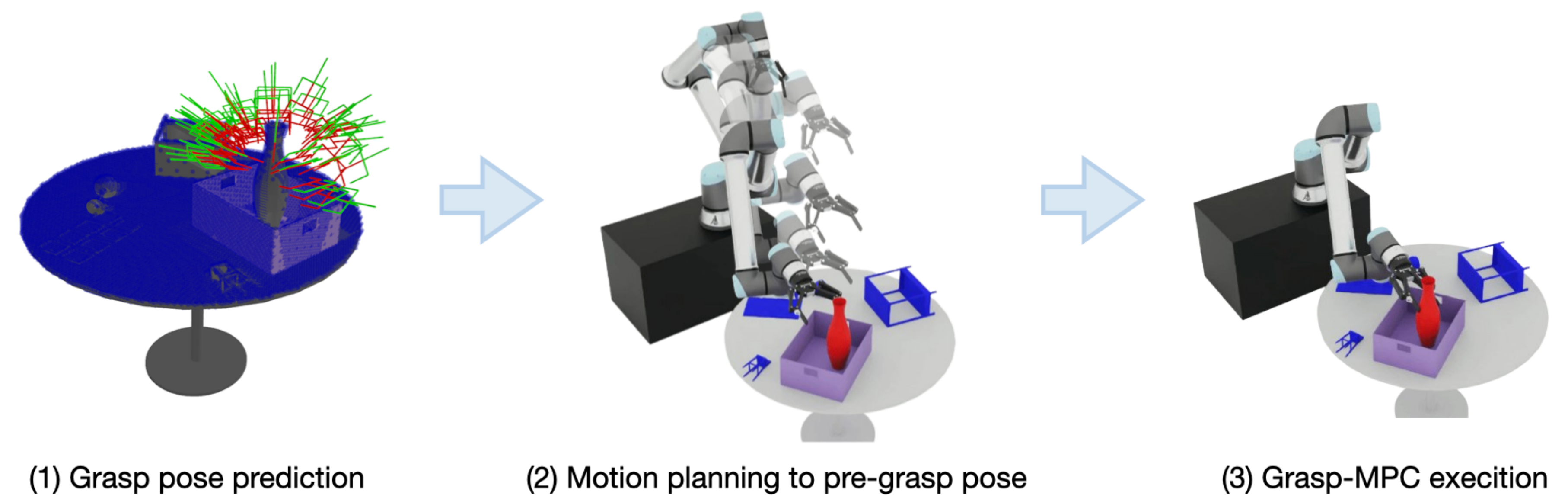}
    \caption{\ourmethod~Pipeline. 
    \ourmethod~seamlessly integrates off-the-shelf grasp prediction and motion planning with an MPC that incorporates a learned grasp value function, enabling grasping in cluttered scenes.
    The pipeline involves: (1) predicting grasp and pre-grasp poses using a fixed offset and filtering out in-collision poses via IK; (2) planning a trajectory to a collision-free pre-grasp pose; and (3) executing actions from the pre-grasp to grasp the object.
    }
    \label{fig:pipeline}
    \vspace{-0.5cm}
\end{figure*}

\ourmethod~leverages a large-scale grasp trajectory dataset generated in a simulation (Section~\ref{sec:data_generation}) to train a value function (Section~\ref{sec:value_trainig}).
Then, the learned value function serves as a cost function within MPC (Section~\ref{sec:value_cost}), enabling robust and safe closed-loop grasping that generalizes to novel objects.
At deployment, \ourmethod~is integrated with a grasp pose prediction model and motion planning to operate in cluttered environments (Section~\ref{sec:deployment}).
Lastly, the implementation details of the value function and MPC are described in Section~\ref{sec:implementation}.

\subsection{Data Generation}
\label{sec:data_generation}
A diverse set of grasp trajectories is generated using $8515$ Objaverse objects~\cite{deitke2024objaverse} (see Figure~\ref{fig:teaser}).
\ourmethod~operates from pre-grasp poses, estimated using an off-the-shelf grasp pose prediction model (Figure~\ref{fig:pipeline} (1)), which provides a rough estimate of viable grasp poses.
Thus, each grasping trajectory is generated to move from a pre-grasp pose to the corresponding ground-truth grasp pose.
The grasp pose annotations are from the GraspGen ~\cite{murali2025graspgen} dataset. 
These poses, generated via antipodal sampling similar to ACRONYM~\cite{eppner2021acronym}, include both feasible and infeasible grasp poses, which are verified for physical feasibility in Isaac Sim.
Pre-grasp poses are derived by applying a fixed $15\text{cm}$ offset from each annotated grasp pose.
To increase data coverage, we add random translation noise sampled from $\mathcal{U}(-0.04\text{cm}, 0.04\text{cm})$ and orientation noise from $\mathcal{U}(-0.04\pi, 0.04\pi)$.

We generate motions from the perturbed pre-grasp poses to the grasp poses using motion planning with CuRobo~\cite{sundaralingam2023curobo}. 
Trajectories that reach physically feasible grasp poses are labeled as successful, and leveraging both successful and failed cases enables the value function to learn grasp success likelihoods.
We do not validate these trajectories with simulation to accelerate data collection.
Up to $256$ trajectories are collected per object, with early termination if motion planning repeatedly fails.
Each sample includes object poses $T^{obj}_{world}$ and end-effector poses $T^{EE}_{world}$.
In total, we collect $2,105,296$ trajectories ($115M$ data points), averaging $55$ steps per trajectory, with lengths ranging from $31$ to $233$ steps.
$70.2\%$ of the collected trajectories reach a successful grasp, and the remaining trajectories are labeled as failures.

\begin{figure}[htb]
    \centering
    \begin{tabular}{c c}
        \includegraphics[width=0.25\linewidth]{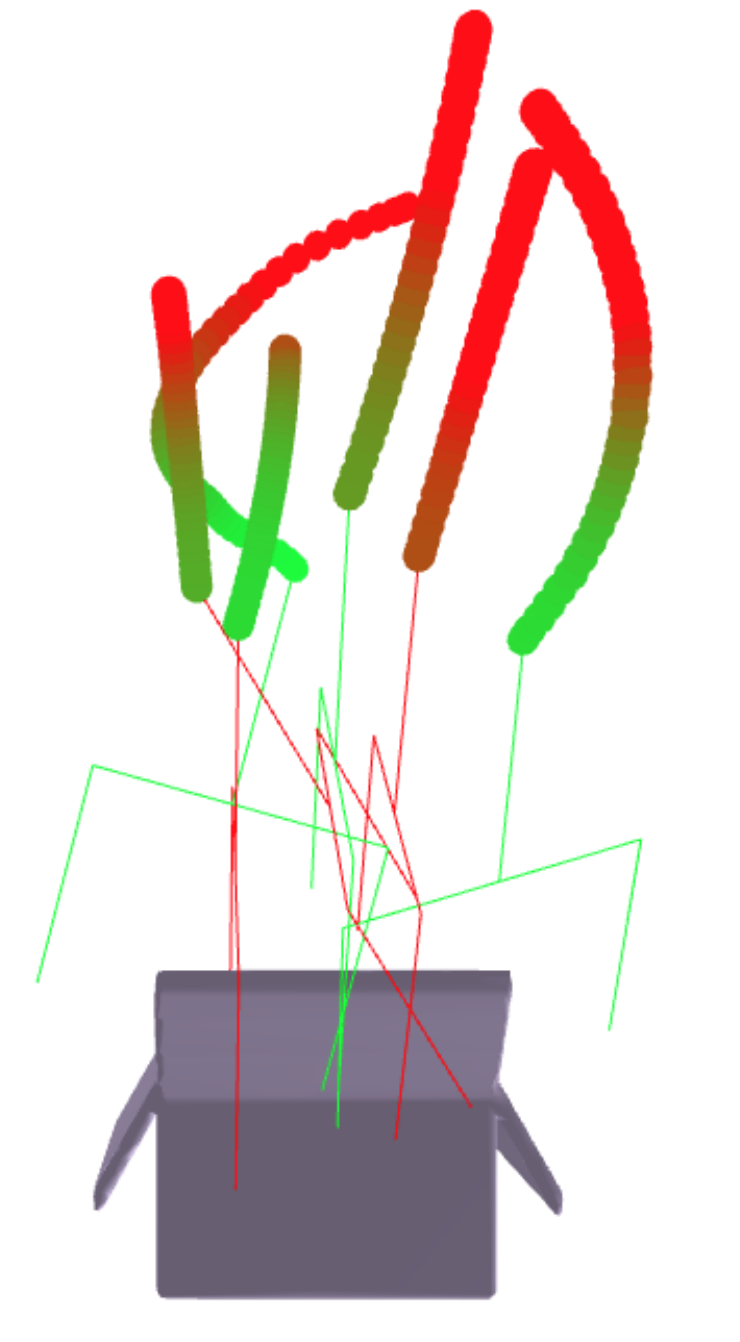}
& 
\includegraphics[width=0.4\linewidth, trim={1cm 0 0 0}, clip]{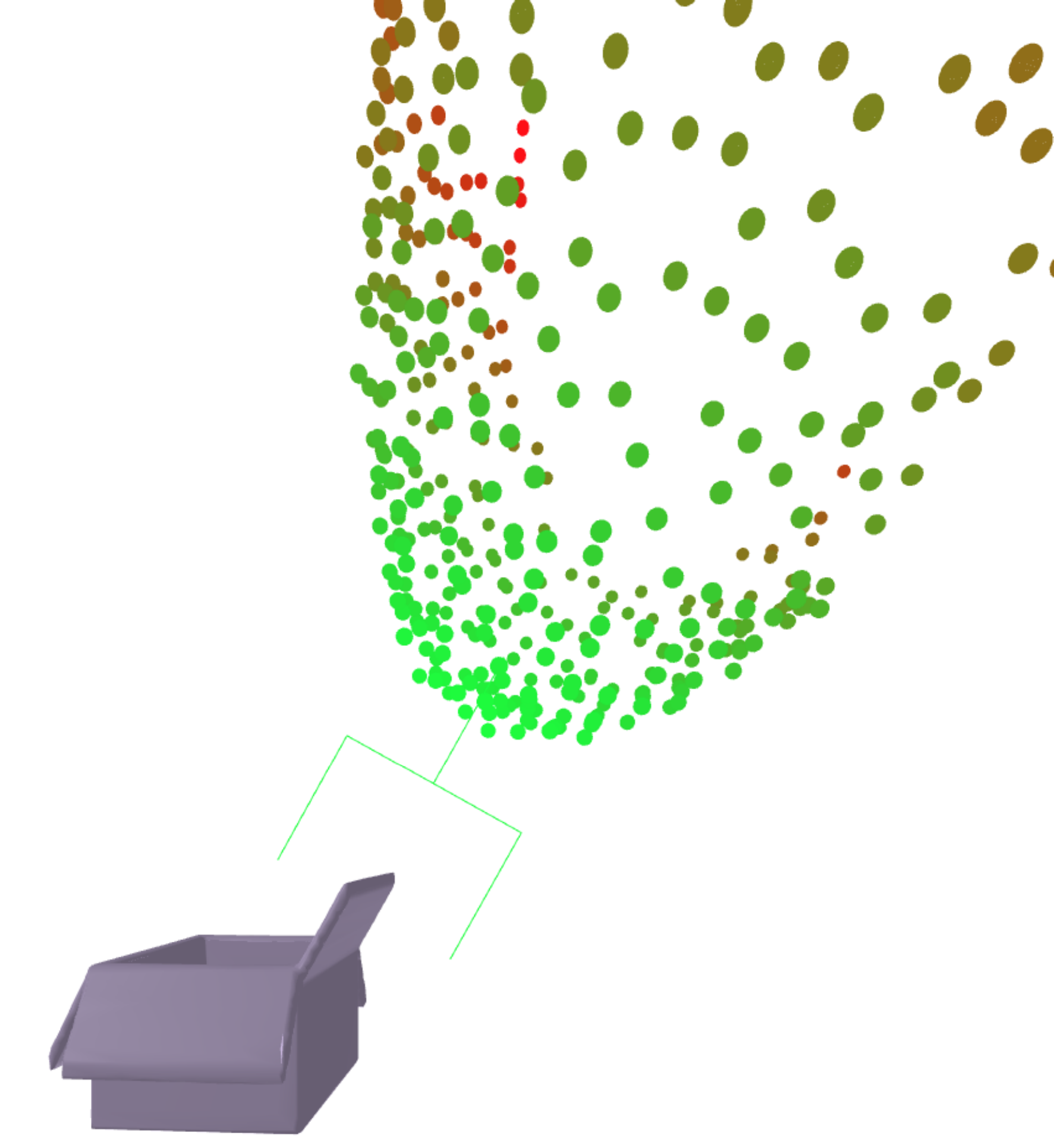}
\\ 
(a) & (b)
    \end{tabular}
    \caption{Visualization of a learned value function. (a) Costs along the collected trajectories for both feasible (\textcolor{green}{green}) and infeasible (\textcolor{red}{red}) grasp poses. The costs at the end of the trajectories for the infeasible grasp poses are higher (darker green) than those for the feasible grasp poses. (b) Estimated values around a desired grasp pose. We keep the orientation of the desired grasp pose and vary $x$ and $y$ coordinates to estimate the values around the grasp pose. \textcolor{red}{Red} points represent higher values (higher in $z$-axis) and \textcolor{green}{green} represents lower values (lower in $z$-axis).}
    \label{fig:value_map}
    \vspace{-0.3cm}
\end{figure}

\subsection{Value Function Training for Grasping}
\label{sec:value_trainig}

\ourmethod~uses a value function as a cost for guiding MPC in grasping.
The value function takes as input~$\mathbf{x}$ consisting of a segmented object point cloud and the end-effector pose relative to the point cloud centroid, $T^{EE}_{obj}$.
To standardize inputs, we center the point cloud by subtracting its mean.
This setup allows \ourmethod~to generalize across the workspace using only local information.
The collected trajectories are labeled with sparse costs, with terminal and near-terminal states in successful grasp trajectories labeled as~$0$, and all others as~$1$.
In particular, the cost $c_{t}$ as timestep $t$ is defined as:
\begin{equation}
\label{eq:cost}
    c_{t} = 
    \begin{cases}
        0 & |q_{goal, i} - q_{t,i}| \leq 5e^{-3}, \ \forall i, \ and \ \mathbbm{1}_{feasible}=1, \\
        1 & \text{Otherwise}
    \end{cases}
\end{equation}
where $q_{t,i}$ and $q_{goal,i}$ are the $i$-th joint positions at time $t$ and the goal, respectively, and $\mathbbm{1}_{feasible}$ indicates whether the trajectory corresponds to a feasible grasp.

A value function $V(\mathbf{x}_{t})$ is then trained to approximate the expected cost-to-go, defined as $V(\mathbf{x}_{t}) = \mathbb{E}_{\tau}[J(\tau)]$.
The value function is trained using the Bellman error objective~\cite{Sutton1998}:

\begin{align}
    y_t &= c_t + \gamma V_{\phi'}(\mathbf{x}_{t+1}) \\
    \ell(\phi; \mathbf{x}_{t}, c, \mathbf{x}_{t+1}) &= \big(y_t - V_\phi(\mathbf{x}_{t})\big)^2 \\
    \phi^* &= \argmin_{\phi}\; \mathbb{E}_{(\mathbf{x}_{t}, c, \mathbf{x}_{t+1})} \left[ \ell(\phi; \mathbf{x}_{t}, c, \mathbf{x}_{t+1}) \right]
\end{align}
where $y_t$ is the one-step target consisting of the immediate cost $c_t$ and the discounted value of the next state~$V_{\phi'}(\mathbf{x}_{t+1})$ from the target value function with exponential moving average of parameters $\phi$.
We set the discount factor to $\gamma = 0.99$, and the exponential moving average uses a rate of $5 \times 10^{-3}$.  
Figure~\ref{fig:value_map} presents a visualization of the learned value function.
The estimated value (cost) is lower (depicted in light green) around the target grasp pose, facilitating MPC in guiding the robot toward a successful grasp pose.

\subsection{Integrating a Value Function as a Grasp Cost within MPC}
\label{sec:value_cost}
\ourmethod~uses learned value function as costs to guide MPC in minimizing grasping cost during online deployment.
The value function approximates the expected cost-to-go, which are integrated into the MPC objective to select control inputs.
However, MPC may sample out-of-distribution actions, leading to unreliable cost estimates and reduced performance. To mitigate this, prior work~\cite{jawale2024dynamicnonprehensileobjecttransport, pmlr-v162-cheng22b, kumar2020conservative} constrains MPC using pessimistic upper bounds derived from ensembles to avoid unsupported states.
In our setting, ensembles do not yield performance improvements, as the large-scale synthetic data already covers a wide distribution of states.
We therefore leave the integration of such risk-averse objectives for future work.
Thus, in this work, the following objective is employed:
\begin{equation}
\label{eq:pess_cost}
    C_{grasp}(\mathbf{x}_{h \in H}) = \Sigma^{t+H}_{t'=t}\gamma^{t'-t} V_{\theta}(\mathbf{x}_{t'})
\end{equation}

\ourmethod~augments standard MPC costs (e.g., minimum jerk, collision) with a value-based grasp cost.
We integrate the grasp cost into CuRobo~\cite{sundaralingam2023curobo}, and define the final cost as:
\begin{equation}
\label{eq:all_cost}
    C_{\ourmethod} = C_{curobo} + \omega C_{grasp}
\end{equation}
where $C_{curobo}$ is a set of CuRobo’s default costs. $\omega$ is a weight for the pessimistic cost function, which we set to 1000.

\subsection{Grasp-MPC Deployment}
\label{sec:deployment}
At deployment, \ourmethod~is combined with an open-loop grasp pose prediction model~\cite{yuanm2t2} and motion planner~\cite{sundaralingam2023curobo}.
The grasp prediction model generates grasp poses for the target object. 
Pre-grasp poses are obtained by moving a negative distance along the gripper's approach vector. 
For real-world experiments, the standard $10$cm offset distance~\cite{yuanm2t2, fang2023anygrasp} is used. 
For simulation experiments in Fetchbench~\cite{han2024fetchbench}, $6$cm offset is used, reduced from the standard $10$cm due to insufficient clearance in many benchmark scenes, but increased from the original paper's $4$cm which is too restrictive. 
Feasible, collision-free pre-grasp poses are verified via a goalset motion plan, and the robot moves to the selected pre-grasp pose using the motion planner.
Once positioned, \ourmethod~grasps the object with feedback of the robot state, segmented object pointcloud, and the world signed distance field.

\subsection{Implementation}
\label{sec:implementation}

While \ourmethod~is compatible with any sampling-based MPC library, we specifically employ Model Predictive Path Integral (MPPI)~\cite{williams2017mppi} implemented in CuRobo\cite{sundaralingam2023curobo}, a GPU-accelerated MPC framework, to enable fast and efficient real-time control.
The value function consists of a PointNet++~\cite{qi2017pointnetdeephierarchicalfeature} for point cloud input and an MLP for proprioception.
Their outputs are concatenated and passed to an MLP head to output a value.
A \textit{softplus} activation ensures positive value predictions.
Training uses both full and partial point clouds, with partial views rendered from randomly placed virtual cameras. 
Gaussian noise is added for robustness to real-world noise.
The value function is trained with a mini-batch size of $1536$, where $32$ distinct object point clouds are sampled and $48$ states are sampled for each object point cloud.
The training procedure was conducted on a single RTX 4090 GPU for a duration of six days, using a learning rate of $1 \times 10^{-4}$.

\section{Experimental Results: Simulation}
\label{sec:experimental_results_sim}
\begin{figure}
    \centering
    \includegraphics[width=1.0\linewidth]{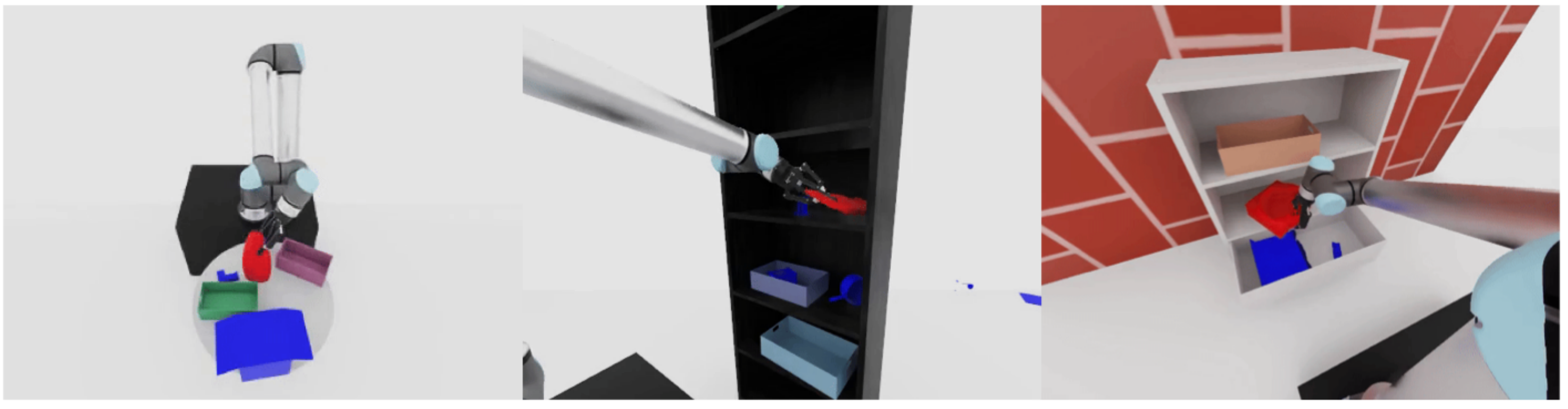}
    \caption{Simulated environments. \ourmethod~is extensively evaluated in FetchBench~\cite{han2024fetchbench} environments.}
    \label{fig:sim_exp_scene}
    \vspace{-0.5cm}
\end{figure}

\begin{figure*}
    \centering
    \includegraphics[width=0.85\linewidth]{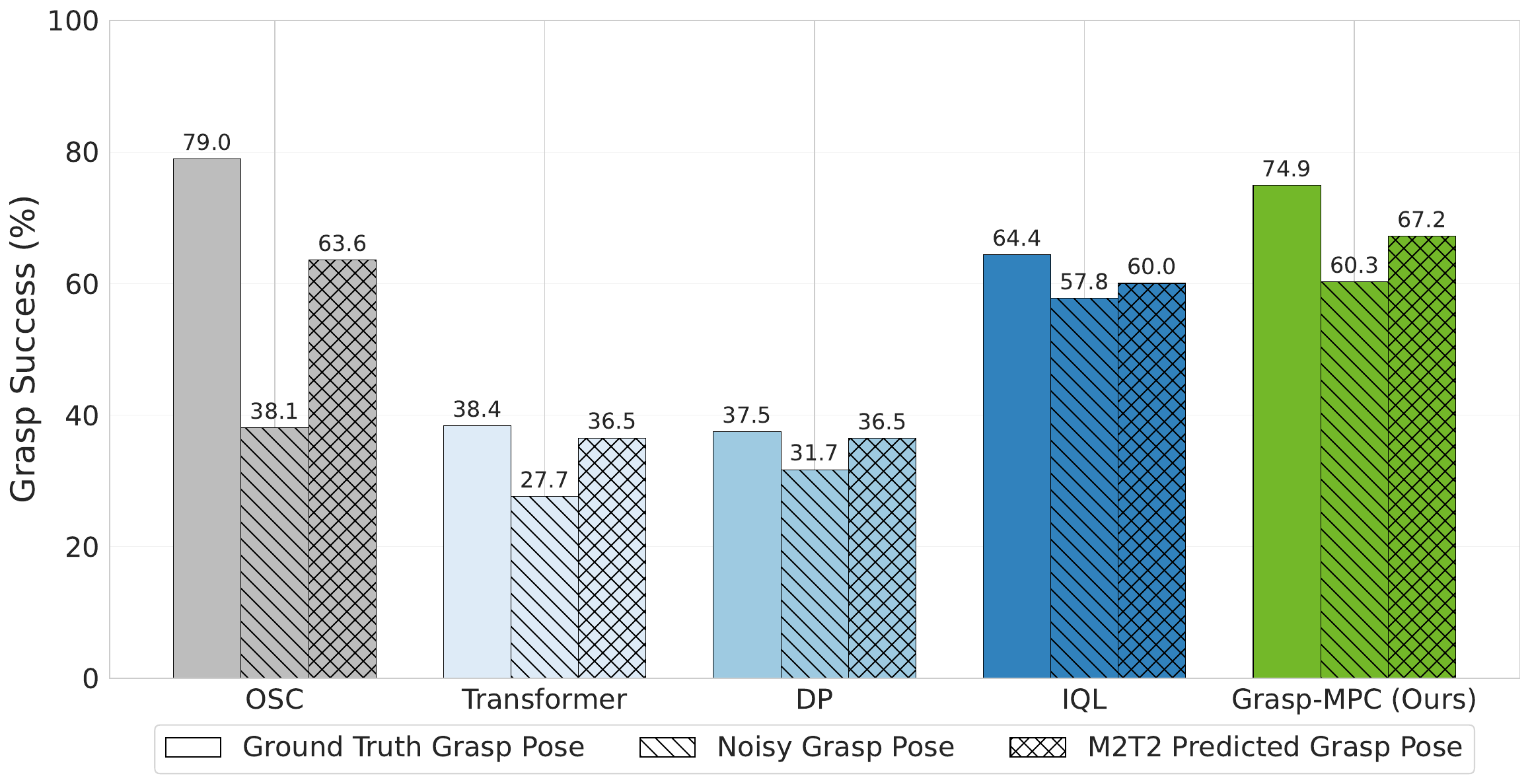}
    \caption{
Grasp performance comparison across different methods given ground truth, noisy, and M2T2-predicted grasp poses. Solid bars represent performance with ground truth grasp poses, diagonally hatched (///) bars show results with noisy perturbations~($\pm2$cm , $\pm18$deg), and cross-hatched (xx) bars indicate M2T2-predicted grasp poses. While \ourmethod~maintains high grasp success rate across all perturbations, OSC, an open-loop baseline, significantly drops with perturbed grasp poses.
    }
    \label{fig:result_sim}
\end{figure*}

In this section, we evaluate \ourmethod~and competitive baselines in simulated environments designed to test grasping capabilities in cluttered settings.
The experiments address: (1) How well does \ourmethod~grasp unseen objects using ground-truth grasp poses? (2) How robust is it to perturbed ground-truth poses? (3) How does it perform with predicted grasp poses?
\subsection{Experiment Setup} 
\ourmethod~and baselines are evaluated in simulation using a UR10 robot with a Robotiq 2F-140 gripper.
Evaluations are based on FetchBench~\cite{han2024fetchbench} (Figure ~\ref{fig:sim_exp_scene}), adapted to use Isaac Sim for improved physics and closed-loop gripper modeling.
All objects are novel and consist of both procedurally generated and the ACRONYM objects~\cite{eppner2021acronym}.
Three cameras capture point clouds, and the one with the greatest coverage of the target object is selected to provide observations to the policy.
Experiments span $90$ scenes ($60$ problems each), resulting in $5,400$ test cases.
A $6cm$ offset is used to compute pre-grasp poses from the grasp poses.

\noindent \textbf{Evaluation Metrics:}
The evaluation uses the metric \emph{Grasp Success}, defined as lifting the object at least $1\text{cm}$. Since the task involves motion planning to reach the pre-grasp pose, this metric accounts for motion planning failures by excluding trials where moving to the pre-grasp pose failed.

\noindent\textbf{Baselines:}
\ourmethod~is compared to the following:
\begin{itemize}

    \item \textbf{OSC}: Similar to the baseline used in Fetchbench, Operational Space Control~\cite{osc} is used to move the end-effector linearly from the pre-grasp to the grasp pose.

    \item \textbf{Transformer Policy} A Transformer-based policy, inspired by the architecture used in OPTIMUS~\cite{dalal2023imitatingtaskmotionplanning}, and included as a baseline in FetchBench~\cite{han2024fetchbench}.

    \item \textbf{Diffusion Policy (DP)}: A diffusion policy~\cite{chi2023diffusion} that takes point cloud observations as input, which is considered one of the state-of-the-art IL approaches.
    
    \item \textbf{IQL}: A policy trained with Implicit Q-Learning (IQL)~\cite{kostrikov2021offline}, a state-of-the-art offline RL method.
\end{itemize}

All methods, including \ourmethod, use CuRobo's motion planner to generate a trajectory from the robot’s initial position to a pre-grasp pose.
The robot then attempts to grasp the object using one of the above methods.
IL policies are trained only on successful trajectories.

\subsection{Grasp Execution with Grasp Pose Annotations}
\label{sec:result_ga}

Grasp pose annotations for objects in the Fetchbench environments were generated for the Robotiq 2f-140 following ACRONYM~\cite{eppner2021acronym}.
Up to 200 kinematically valid, collision-free grasp poses are randomly sampled during evaluation, from which one is selected for the robot to move to the corresponding pre-grasp pose using a goalset motion planner in CuRobo~\cite{sundaralingam2023curobo}.

\textbf{\ourmethod~nearly matches oracle performance while surpassing closed-loop baselines.}
As shown in Figure~\ref{fig:result_sim}, \ourmethod~demonstrates competitive performance compared to \emph{OSC}, an open-loop approach, which serves as an oracle baseline.
\ourmethod~achieves a grasp success rate of $74.9\%$, closely matching the oracle baseline \emph{OSC} at $79.0\%$.
Moreover, \ourmethod~significantly outperforms closed-loop baseline methods.

\textbf{Closed-loop baselines underperform due to suboptimal data and domain mismatch.}
IQL achieves a grasp success of $64.4\%$, below \ourmethod~(73.6\%), likely due to limitations in its policy extraction process~\cite{park2024value}.
IL-based approaches also perform poorly, likely because motion planning, while efficient for data collection, often yields suboptimal demonstrations.
Additionally, the gap between the empty scene used for data collection and the object rich evaluation environments introduce MDP mismatches that further degrade performance.

\textbf{\ourmethod~is robust to noisy pre-grasp poses.}
\ourmethod~and baseline methods are also evaluated using ground-truth grasp poses perturbed with random noise sampled from $\mathcal{U}(-2\text{cm}, 2\text{cm})$ in translation and $\mathcal{U}(-18\text{deg}, 18\text{deg})$ in orientation.
As shown in Figure~\ref{fig:result_sim}, \emph{OSC} fails to recover from these perturbations due to its open-loop nature, resulting in a 40\% drop in performance. 
In contrast, \ourmethod~achieves a $60.3\%$ grasp success rate (14\% drop), outperforming baseline closed-loop control policies.

\subsection{Grasp Execution with a Grasp Pose Prediction Model}

M2T2~\cite{yuanm2t2} is used as an off-the-shelf grasp prediction model to generate target grasp poses.
We attempted to train M2T2 using grasp pose annotations specific to the Robotiq gripper, but could not train an optimal grasp pose prediction model.
We instead use the publicly available M2T2 model trained on Franka Panda gripper, adding a $10 \text{cm}$ offset to adapt the predicted grasp poses for the Robotiq gripper.
The top-$512$ scored grasp poses are selected.

\textbf{\ourmethod~maintains strong performance despite noisy grasp poses from M2T2, outperforming all baselines.} 
IL-based approaches perform poorly achieving only 36.5\% grasp success rate as shown in Figure~\ref{fig:result_sim}. The standard OSC approach achieves 63.6\% success rate, dropping by 15\% from using ground truth grasp poses. \ourmethod~achieves 67.2\% success, highest among methods, dropping only by 8\% from using ground truth grasp poses. 
This result highlights \ourmethod's robustness to prediction errors from an off-the-shelf grasp prediction model and shows promise for real world deployment which is evaluated in the next section.

\begin{figure*}[t]
    \centering
    \includegraphics[width=\linewidth]{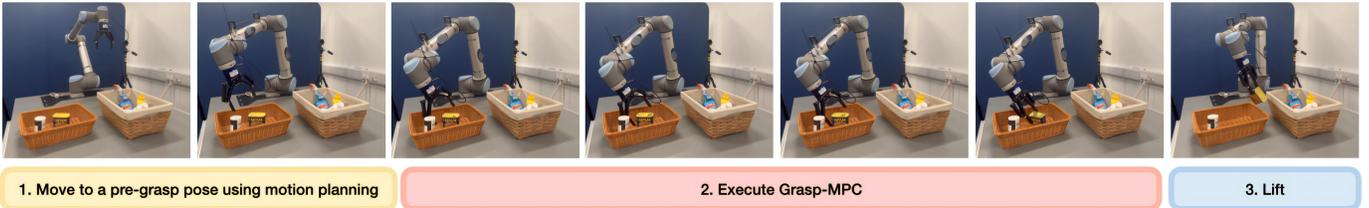}
    \caption{\ourmethod~execution in the Table Clutter scene. \ourmethod~effectively grasps a novel object from the bin. }
    \label{fig:real_rollout}
    \vspace{-0.3cm}
\end{figure*}

\section{Experimental Results: Real-World}

\begin{figure}%
    \centering
    \includegraphics[width=1.0\linewidth]{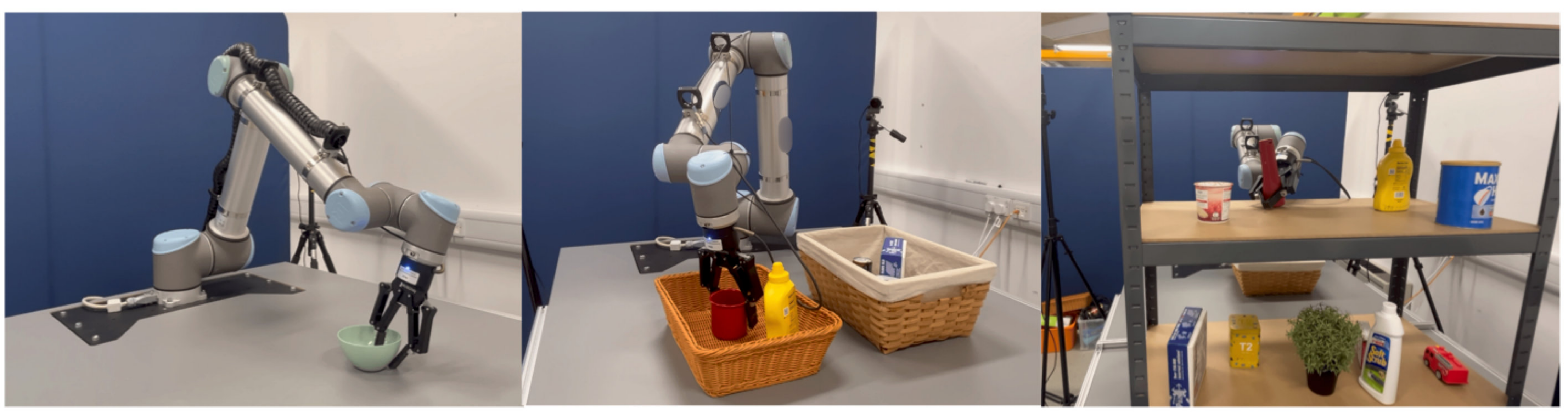}
    \caption{Representative real-world environments: (Left) Table Empty, (Middle) Table Clutter, and (Right) Shelf Clutter.}
    \label{fig:real_env}
    \vspace{-0.3cm}
\end{figure}

\ourmethod~is evaluated on novel objects using a UR10 robot arm with a Robotiq 2F-140 in real-world environments.
The experiments are designed to address:
(1) How effectively does \ourmethod~grasp novel objects in challenging real-world environments compared to open-loop approaches?
(2) How well does it adapt to dynamic perturbations when the target object pose changes after reaching the pre-grasp position?

\subsection{Experiment Setup}
\textbf{Perception System.} 
Two RealSense L515 cameras with known extrinsics capture $640\times480$ depth images to generate point cloud observations.
Target object segmentation is performed using SAM-Track~\cite{cheng2023segmenttrack}, which combines Grounding DINO~\cite{liu2023grounding} for detection and SAM~\cite{kirillov2023segment} for segmentation, producing input point clouds for \ourmethod's value function.
To handle obstacles in cluttered scenes, NVBlox~\cite{millane2024nvbloxgpuacceleratedincrementalsigned} is used to represent the environment for CuRobo's motion planning and MPC modules.

\textbf{Success Criteria.} 
Grasp success is defined as lifting the target object and returning the arm to its home position without dropping it during execution.

\textbf{Grasp Pose Prediction.} 
As in the simulated experiments, M2T2 predicts grasp poses, with a $10\text{cm}$ offset applied to adapt them for the Robotiq gripper.

\textbf{Baselines.}
\emph{CuRobo-GraspAPI}, an open-loop approach, is used for comparison, as it provides a simple and reliable grasp execution pipeline suitable for real-world deployment. 
Other closed-loop control policies are excluded from real-world experiments due to safety concerns, as they lack collision avoidance mechanisms and are prone to collide with surrounding obstacles such as a shelf or clutter. 
In contrast, \ourmethod~minimizes collision and learned grasp task costs, enabling safe and effective operation in cluttered real-world environments.

\subsection{Grasping Performance Across Different Scenes}
\textbf{Comprehensive evaluation across three environments with increasing complexity.} 
We compare \ourmethod~against CuRobo-GraspAPI across three progressively challenging environments: \emph{empty tabletop}, \emph{cluttered tabletop}, and \emph{cluttered shelf scenes} as shown in Figure~\ref{fig:real_env}. 
Each environment utilizes $5$ different objects, with distinct object sets for each environment to ensure comprehensive evaluation across varying complexity levels. 
For each object, three different poses are considered, and two evaluation trials per method are performed to assess consistency and reliability.
In total, $30$ trials are conducted for each environment.

\begin{figure}
    \centering
    \includegraphics[width=0.9\linewidth]{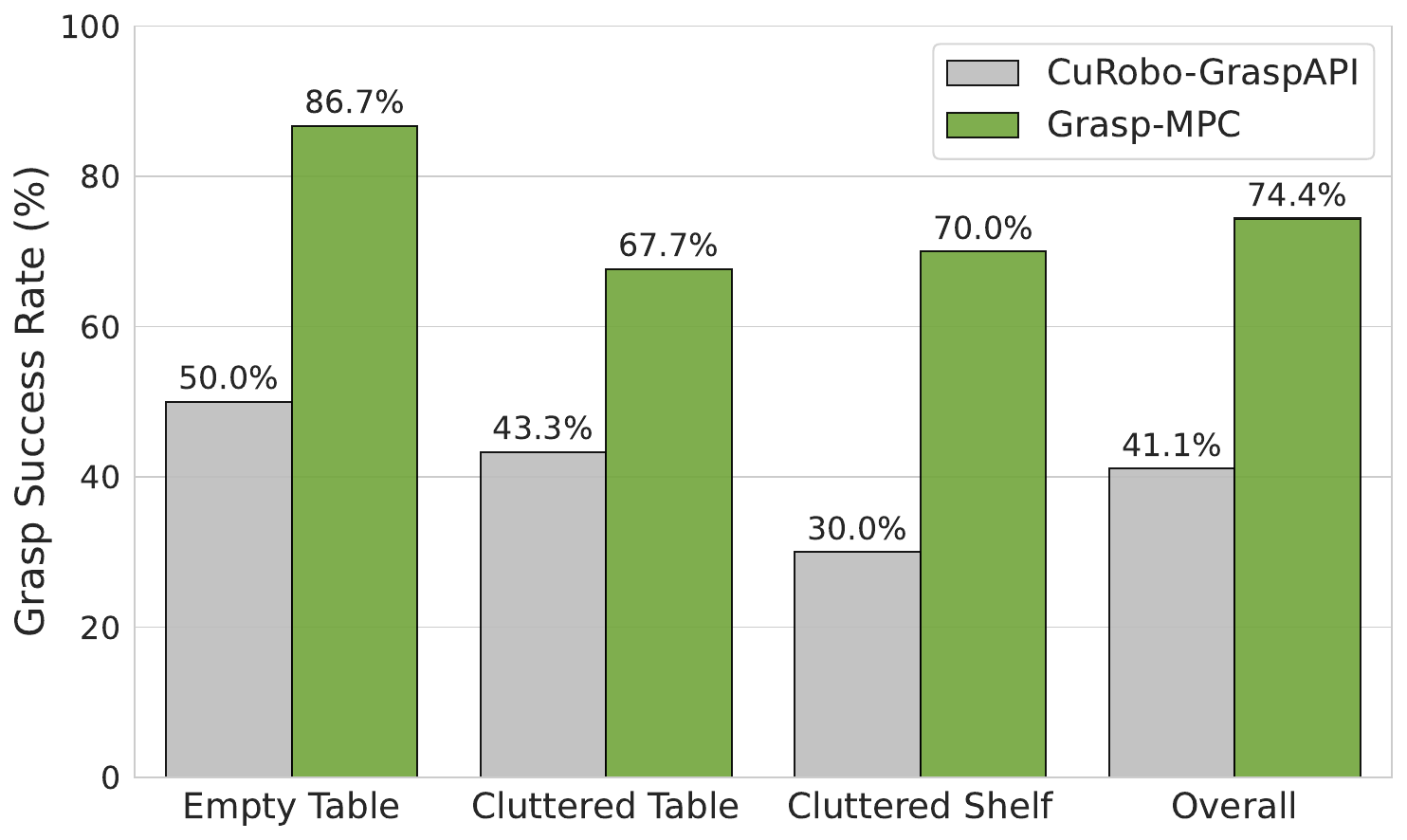}
    \caption{Grasp performance comparison between Open-loop (CuRobo-GraspAPI) and \ourmethod. \ourmethod~consistently achieves higher grasp success rates across all scenes, highlighting its robustness in real-world environments.}
    \label{fig:grasp_comparison}
\end{figure}

\textbf{Consistent outperformance over open-loop baselines across all environments.} 
Figure~\ref{fig:grasp_comparison} presents the grasp success rates of \ourmethod~and CuRobo-GraspAPI across different scenes. 
The open-loop baseline frequently fails to grasp target objects when the predicted grasp pose deviates from the ideal configuration, as it cannot adapt during execution. 
In contrast, \ourmethod~continuously adjusts the gripper pose to minimize the task grasp cost function while avoiding obstacles such as a shelf, demonstrating safe and robust grasp execution across all environments. 
\ourmethod~outperforms the baseline in all scene types, with greater performance improvements observed in complex environments.

\begin{figure*}[t]
    \centering
    \includegraphics[width=\linewidth]{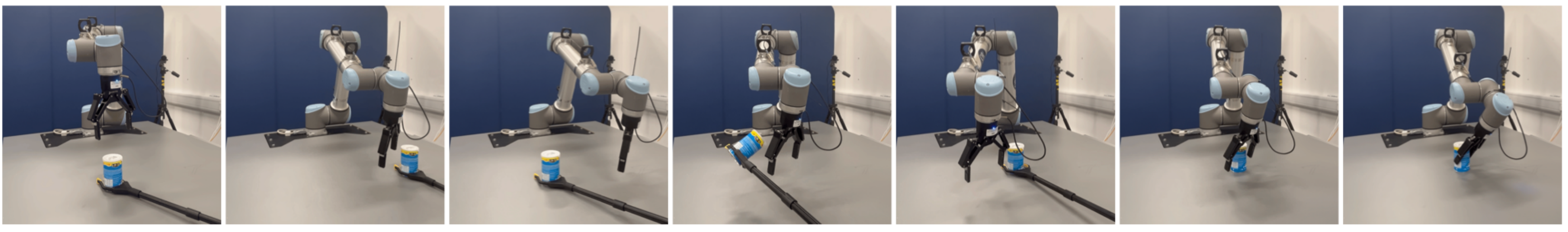}
    \caption{\ourmethod~execution for moving objects. \ourmethod~adapts in real time to track and grasp moving target objects, capabilities that open-loop approaches lack.}
    \label{fig:moving_object}
    \vspace{-0.5cm}
\end{figure*}

\subsection{Real-Time Adaptation to Object Pose Perturbations}
To demonstrate the benefits of \ourmethod~as a closed-loop control policy, perturbations are added to the target object pose after the robot reaches the pre-grasp pose. 
Since open-loop approaches cannot handle such dynamic changes by design, we evaluate only \ourmethod~in this experiment to assess its real-time adaptation capabilities.

\textbf{Successful grasping despite large object perturbations during execution.}
We evaluate $5$ objects with $6$ trials per object, resulting in $30$ trials total. 
Figure~\ref{fig:moving_object} illustrates an example rollout demonstrating \ourmethod's adaptability. 
\ourmethod~achieves a $60\%$ success rate, demonstrating its capability to adapt in real-time even when large perturbations are applied to the target object pose. 
\ourmethod~exhibits this global behavior while having trained on grasping motions that are only $15$cm away from a grasp. 
We are planning on training with larger grasping motions in the future to study if we can get even higher success at this task.

\section{Discussion}

This work presents \ourmethod, a 6-DoF closed-loop grasping policy for novel objects in cluttered environments.
The approach learns a value function from both successful and failed grasping trajectories, which is subsequently integrated into an MPC framework to generate actions during deployment.
Through its modular design, \ourmethod~can address deployment-time constraints such as clutter without requiring retraining.
Moreover, leveraging both successful and failed trajectories enhances data efficiency.

\ourmethod~is validated on the simulated benchmark FetchBench~\cite{han2024fetchbench} across 5,400 grasping problems, demonstrating superior performance compared to imitation learning methods.
It also outperforms classical planning-based approaches in scenarios with noise in the grasp pose or when the grasp pose was provided by a learned model (e.g., M2T2).
On a real robotic system, \ourmethod~achieves a $30\%$ higher success rate than a planning-based approach in cluttered table-top and shelf settings, despite being trained exclusively on empty scenes.
These results demonstrate the ability of the method to operate with noisy point clouds and to handle real-world contact physics without relying on physically simulated training data.

While shown to be effective, \ourmethod~presents several limitations that highlight opportunities for future improvement and extension.
First, although higher success rates are achieved compared to existing methods, performance in real-world deployments does not yet reach $100\%$.
We hypothesize that leveraging physics simulation during data generation to generate success/failure labels will improve the performance of \ourmethod. 
\ourmethod~can also be readily finetuned with real-world data as only success/failure labels are required for trajectories. 
Both of these explorations are left to future work.

A further limitation is that validation is restricted to grasping tasks.
The approach could, in principle, extend to other manipulation tasks if demonstrations can be generated using existing pipelines~\cite{garrett2024skillmimicgen}.
We leave studying the performance of value function learning across diverse manipulation tasks for future work.

\section*{ACKNOWLEDGMENT}

This work was supported by a UKRI/EPSRC Programme Grant [EP/V000748/1]. We would like to acknowledge the use of the University of Oxford Advanced Research Computing (ARC) and SCAN facilities in carrying out this work. We would like to thank Stan Birchfield for providing feedback on early versions of this paper.

\bibliographystyle{IEEEtran}
\bibliography{bib/conference, bib/grasp, bib/misc, bib/rl, bib/robotics}  

\clearpage
\appendix
\subsection{Full point cloud vs partial point cloud observations.}
\ourmethod~robustly grasps novel objects even when provided with partial point cloud observations, achieving a grasp success rate of $74.9\%$, which is only $1.4\%$ lower than with full point clouds ($76.5\%$). 
This small performance gap highlights its practicality for real-world deployment, where sensor occlusions and incomplete observations are common, and ensures reliable grasping performance despite imperfect perception.

\subsection{Grasp Success by Scene-Type in Simulation}

\begin{figure}
    \centering
    \includegraphics[width=\linewidth]{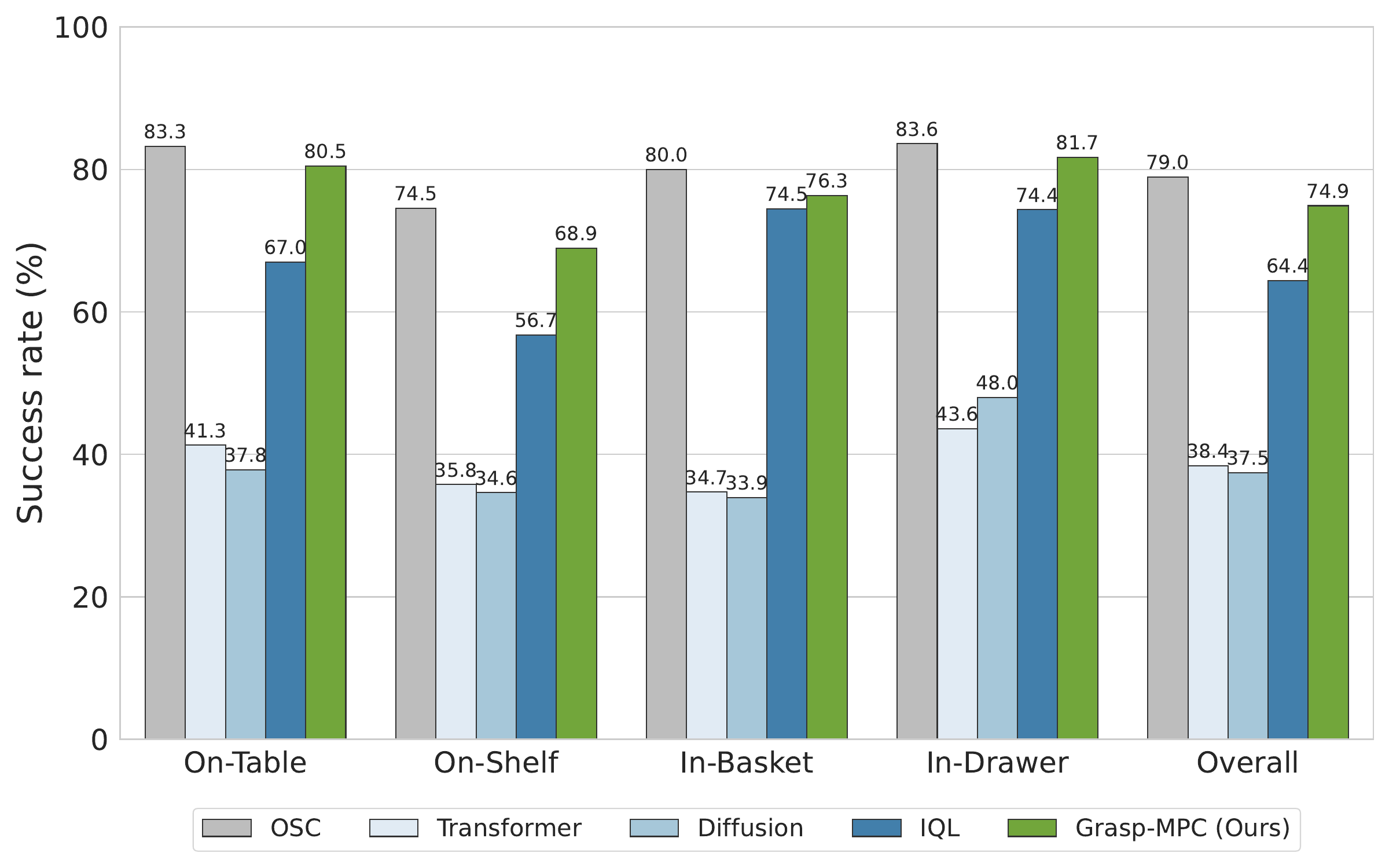}\\
    (a) Ground Truth Grasp Pose Annotations\\
    \includegraphics[width=\linewidth]{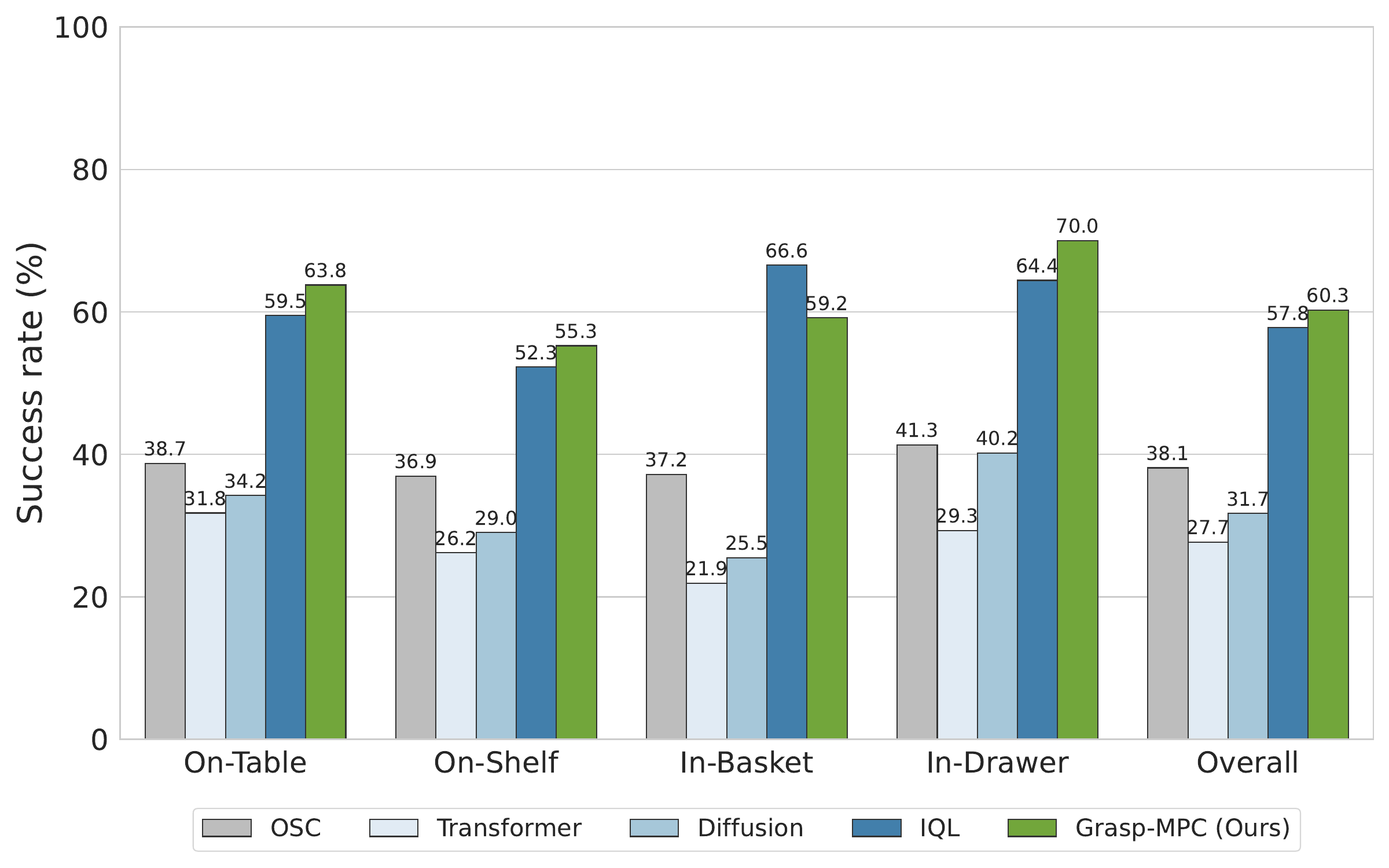} \\
    (b) Noisy Grasp Pose Annotations\\
    \includegraphics[width=\linewidth]{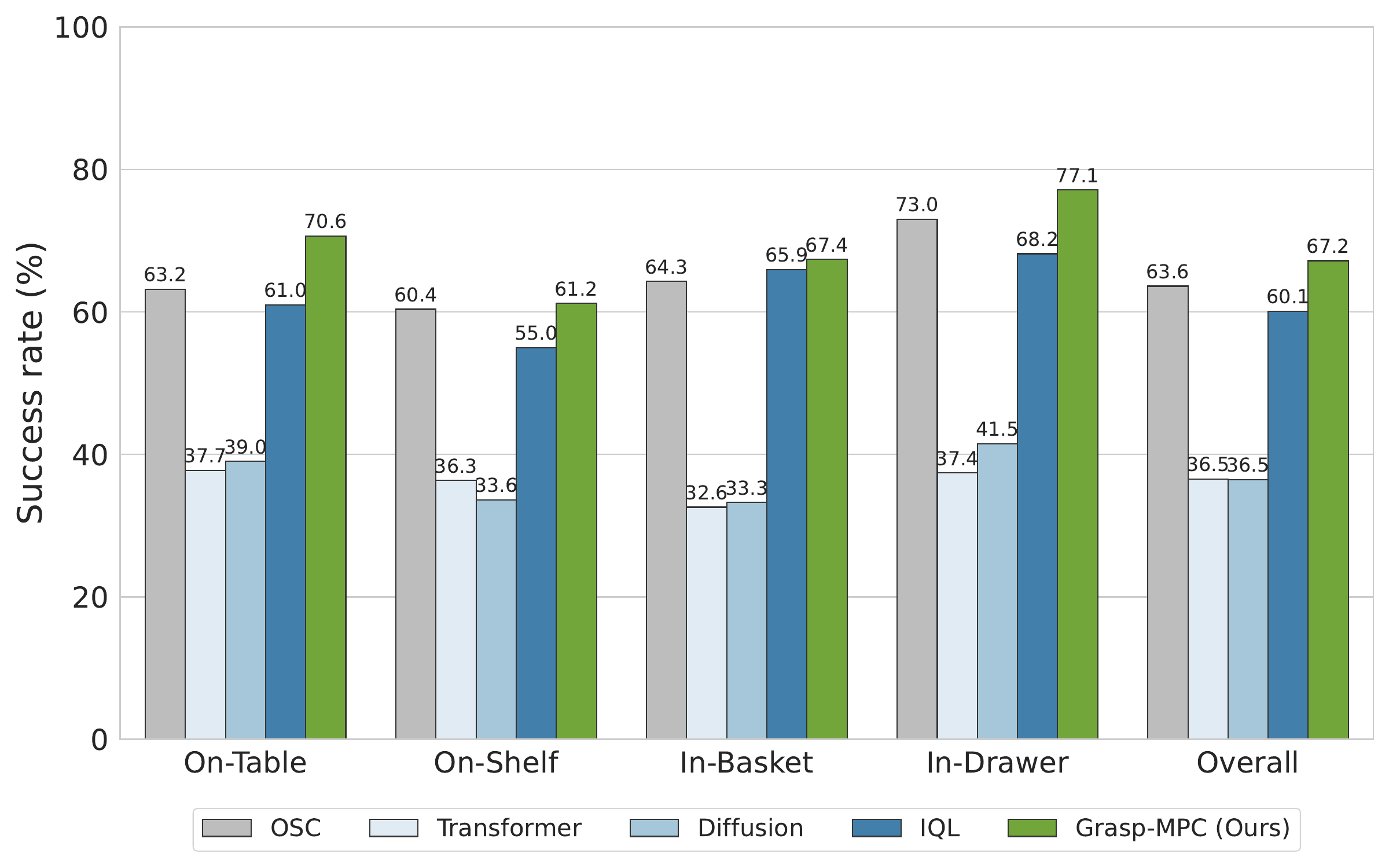} \\
    (c) M2T2 generated Grasp Poses
    \caption{Grasp success rate for each scene type given grasp pose annotations from ground truth (a), added noise (b), and from a trained grasp pose prediction model M2T2 (c). \ourmethod~achieves a competitive or substantially higher success rate compared to the competitive baselines.} 
    \label{fig:grasp_success_category}
\end{figure}

\textbf{\ourmethod~performs competitively with the oracle baseline (\emph{OSC}) and consistently outperforms other baselines across all scene categories; however, \emph{on-shelf} scenes remain particularly challenging.} 
Figure~\ref{fig:grasp_success_category} shows the grasp success rates across various scene categories.
While \emph{OSC} serves as an oracle baseline by leveraging ground-truth grasp poses, \ourmethod~achieves a comparable success rate, demonstrating strong performance.
Moreover, \ourmethod~consistently outperforms all non-oracle baselines across every scene category.
Among all scene categories, \emph{on-shelf} environments are the most challenging, a trend that also holds in real-world settings (see Figure~\ref{fig:grasp_comparison}).

\textbf{\ourmethod~consistently outperforms baseline methods, and the performance of \emph{OSC} drops significantly when noisy grasp poses are given.}
Figure~\ref{fig:grasp_success_category}-(b) presents grasp success rates across different scene categories using grasp poses perturbed with random noise sampled from $\mathcal{U}(-2\text{cm}, 2\text{cm})$ in translation and $\mathcal{U}(-18\text{deg}, 18\text{deg})$ in orientation.
Unlike in the ground-truth setting, the performance of \emph{OSC} degrades substantially under noise, as its open-loop nature prevents it from recovering from perturbations.
In contrast, \ourmethod~maintains strong performance even with noisy grasp inputs.
Interestingly, \emph{IQL} slightly outperforms \ourmethod~in the \emph{In-Basket} scene; however, \ourmethod~achieves consistently superior results across the remaining scenes and in overall performance.
We hypothesize that baseline methods such as \emph{IQL} struggle in more constrained environments like \emph{On-Shelf} and \emph{In-Drawer} scenes.

\textbf{\ourmethod~consistenly has a higher grasp success rate in each scene type when combined with a grasp prediction model.}
Figure~\ref{fig:grasp_success_category}-(c) illustrates the grasp success rate of \ourmethod~and the baseline approaches for each scene type.
\ourmethod~consistently achieves success rates exceeding $60\%$ across all scene categories and outperforms baseline approaches by a significant margin.
This highlights the superior efficacy of \ourmethod~even when target grasp poses are predicted by an off-the-shelf grasp pose prediction model.

\subsection{Ensemble Ablation Study}
We present results using a single value function, as ensembles did not yield meaningful performance improvements. To validate this choice, we use the following risk-averse objective for ensemble value functions, following prior work~\cite{jawale2024dynamicnonprehensileobjecttransport}:

\begin{equation} \label{eq:pessi_cost} C_{grasp}(\mathbf{x}_{h \in H}, \ddot{\mathbf{a}}_{h \in H}) = log(\sum^{K}_{1} exp(\frac{1}{\lambda} G_{i}(\mathbf{x}_{h\in H}, \ddot{\mathbf{a}}_{h\in H}))) \end{equation}

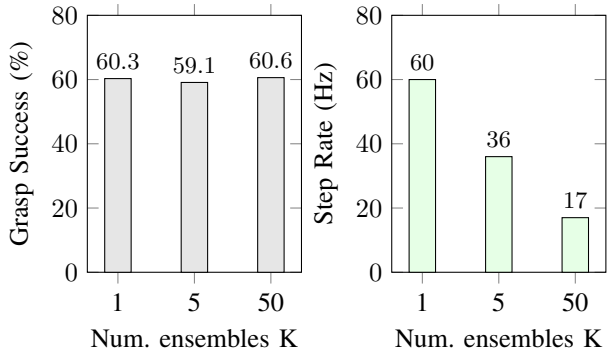
\begin{figure}
\centering
\begin{tikzpicture}
\begin{groupplot}[
    group style={
        group size=2 by 1,
        horizontal sep=1.2cm
    },
    width=0.5\linewidth,
    height=5cm,
    ybar,
    symbolic x coords={1,5,50},
    xtick=data,
    enlarge x limits=0.2,
    nodes near coords,
    every node near coord/.append style={font=\small},
    xlabel={Num. ensembles K},
]

\nextgroupplot[
    ylabel={Grasp Success (\%)},
    ymin=0, ymax=80,
]
\addplot[
    fill=black!10,
    nodes near coords={        
            \pgfmathprintnumber{\pgfplotspointmeta}
}
] coordinates {
    (1,60.3)
    (5,59.1)
    (50,60.6)
};

\nextgroupplot[
    ylabel={Step Rate (Hz)},
    ymin=0, ymax=80]
\addplot[
    fill=green!10,
    nodes near coords={
        \pgfmathprintnumber{\pgfplotspointmeta}
        }] coordinates {
    (1,60)
    (5,36)
    (50,17)
};
\end{groupplot}
\end{tikzpicture}
\caption{Ablation on Ensemble showing Grasp Success and Step Rate for different number of ensembles~$K$.}
\label{fig:ensemble_ablation_side}
\end{figure}

\textbf{Impact of Value Function Ensembles.}
Figure~\ref{fig:ensemble_ablation_side} reports grasp success rates across all scenes.
Using an ensemble improves performance by only $0.3\%$, suggesting that the dataset provides sufficient coverage for training a single value function.
Although ensembles would promote safer behavior via pessimistic cost estimation, a single value function still generalizes well and enables effective grasping.
The MPC step runs at $60 hz$ for $K = 1$ and $17hz$ for $K = 50$, indicating the computational trade-off of using larger ensembles.

\begin{figure}
\centering
\begin{tikzpicture}
\begin{axis}[
    ybar,
    bar width=12pt,
    ymin=0, ymax=100,
    ylabel={Grasp Success (\%)},
    xlabel={Ensemble Disagreement Parameter $\lambda$},
    symbolic x coords={0.1,0.5,1,10,50,100,500},
    xtick=data,
    nodes near coords,
    nodes near coords align={vertical},
    width=\linewidth,
    height=5cm,
    enlarge x limits=0.15,
    every node near coord/.append style={font=\small},
]

\addplot[
    fill=black!10,
    point meta=explicit symbolic
] coordinates {
    (0.1,77.1) [77.1]
    (0.5,79.0) [79.0]
    (1,81.5) [81.5]
    (10, 82.3) [\textbf{82.3}]
    (50,81.5) [81.5]
    (100,80.4) [80.4]
    (500,73.1) [73.1]
};

\end{axis}
\end{tikzpicture}
\caption{Grasp Success across different $\lambda$ values.}
\label{fig:disagreement_barchart}
\end{figure}
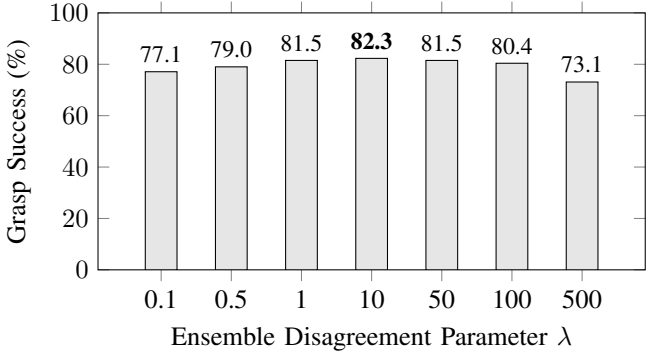

\textbf{Impact of Disagreement Hyperparameter.}
We analyze the effect of the disagreement hyperparameter $\lambda$ in the pessimistic cost when using an ensemble value function with $K=50$.
Figure~\ref{fig:disagreement_barchart} shows grasp success rate of \ourmethod~across $300$ problems with varying $\lambda$.
While performance is generally robust, overly pessimistic costs (i.e., large $\lambda$) can lead to performance degradation.

\subsection{Analyzing Success in FetchBench}

\begin{figure*}
    \centering
    \includegraphics[width=0.9\linewidth]{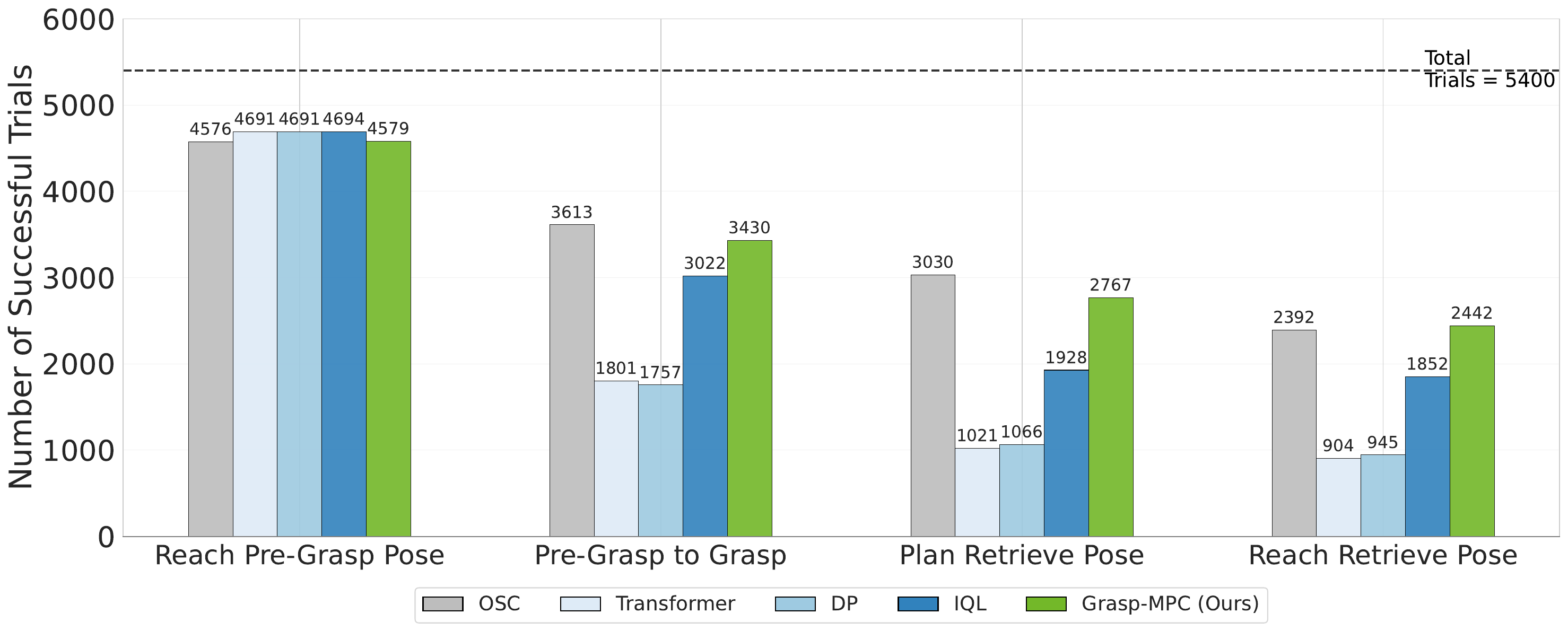}\\
    (a) Ground Truth Grasp Poses\\
      \includegraphics[width=0.9\linewidth]{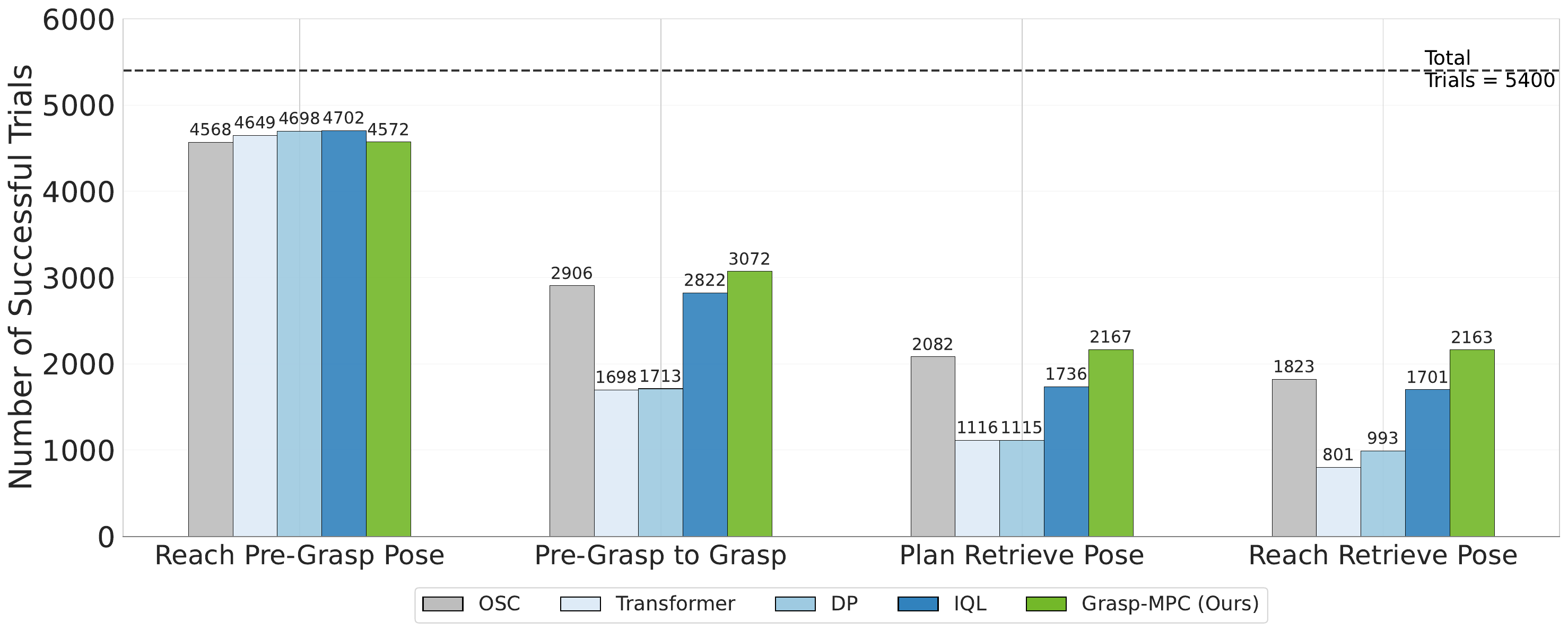}\\
      (b) M2T2 predicted Grasp Poses
    \caption{
         Split of fetchbench task into phases with successful trials at each phase. Each bar represents the number of successful trials for each method across the different phases of task execution in FetchBench. The dashed line indicates the total number of trials (5,400).
    }
    \label{fig:result_fetchbench_task}
\end{figure*}

The Fetchbench benchmark contains 5400 object retrieval problems. The task success, as described by Fetchbench requires the robot to start from an initial joint configuration and grasp a specific object that's in the environment. After grasping the object, the robot then needs to move the grasped object to a retrieve pose. The evaluation done by Fetchbench focuses on the whole object retrieval pipeline while \ourmethod~only solves one part of the pipeline. To make this clear, we split the task into four sequential phases:

\noindent \textbf{Reach Pre-grasp Pose:} The first phase requires the robot to reach a pre-grasp pose. 

\noindent \textbf{Pre-Grasp to Grasp:} Then, from the Pre-Grasp pose, the robot moves to the grasp pose using a linear motion with OSC or with one of the learned policies described in Sec.~\ref{sec:experimental_results_sim}. After closing the gripper on the object, the robot lifts the object by $1$cm with OSC. 

\noindent \textbf{Plan Retrieve Pose:} If the object is still in the gripper after the lift, the motion planner is called to plan a path to reach the retrieve pose. Failures in this phase are motion planning failures. 

\noindent \textbf{Reach Retrieve Pose:} For those problems, where a plan is found, the robot then executes the trajectory. After reaching, if the object is still in the gripper, then task success is achieved. During this phase, failures happen because of the object slipping during motion, leading to either dropping the object or colliding with the environment.

By splitting the task into four sequential phases, failures can be better understood. E.g., even with ground truth annotated grasp poses, the planner was only able to reach the pre-grasp pose 84\% of the problems, as shown in Figure~\ref{fig:result_fetchbench_task}. This could mean that for the remaining problems, a more contact-rich non-prehensile action could be needed to move the object before grasping. Our experiments also showed a 3\% variation in the success of Reach Pre-Grasp Pose between methods (baselines and \ourmethod) even though all methods use the same motion planner. We hypothesize that this could be because of small changes in the initial joint state due to simulation errors. To remove the effect of this slight variation, we calculate grasp success only based on trials that succeeded in Reach Pre-Grasp Pose Phase per method. 

After grasping the object and lifting $1$cm, we observed that not all successful trials in this phase can obtain a collision-free motion plan to move the object to the retrieve pose as seen by the drop in successful trials in \emph{Plan Retrieve Pose} phase in Figure~\ref{fig:result_fetchbench_task}. With ground truth grasp poses, only 83\% of oracle (OSC) and 80\% of \ourmethod~grasped objects get a successful retrieve pose plan. With grasp pose generated by M2T2, the success is worse with only 70\% of OSC and 71\% of \ourmethod~grasped objects getting a successful retrieve motion plan. This highlights the need for grasping methods to also reason about future tasks, like retrieving paths.

Similarly, only 78\% of the planned paths succeed in reaching the retreive pose with the OSC grasped object upon execution while 88\% of the \ourmethod~grasped object reach the retrieve pose when ground truth grasp poses are generated. When using grasps generated from M2T2, the success of reaching the retrieve pose from planned paths increases to 87.5\% for OSC grasped objects and 99.8\% for \ourmethod~grasped objects as shown in Figure~\ref{fig:result_fetchbench_task}-(b). This is promising as even though our training dataset did not contain physics validated (e.g., sim) large retrieve motions to label success/failure. We hypothesize that the learned value function has attempted to go to a region with the most grasps on an object (which often is the most stable grasp). The success analysis across phases in Fetchbench evaluation provides venues for future research, including improvements to the collected dataset. Sim validation of the grasping motion and using large retrieve motions after grasping to label grasp success would improve task success. Conditioning the dataset and value function with retrieve paths will also improve success in Plan Retreive phase.

\subsection{Synthetic Grasp Trajectory Data Generation Details}
\label{appendix:data_collection}
We utilize a motion planner implemented in CuRobo~\cite{sundaralingam2023curobo} to generate trajectories for both valid and invalid grasp poses.
To build the dataset, we spawn $24$ robots at a time, each planning trajectories toward different desired grasp poses for the same object.
Then, we collect up to 256 trajectories for each object.
During data collection, $70\%$ of the trajectories correspond to valid grasp poses, while the remaining $30\%$ represent invalid grasp poses.
The motion planner occasionally struggles to generate viable solutions, leading to delays in the trajectory collection process.
To mitigate this issue and maintain efficiency, we define a maximum failure threshold of 10 attempts per object.
If the failure threshold is reached, we save the trajectories collected so far and reset the collection process to move on to a new object.
This constraint prevents excessive computational overhead during data collection.
This procedure is repeated across a total of $8,151$ objects sourced from the Objaverse dataset.

\subsection{Value Function Training Details}
\label{appendix:value_function}

\begin{figure}[t]
    \centering
    \includegraphics[width=\linewidth]{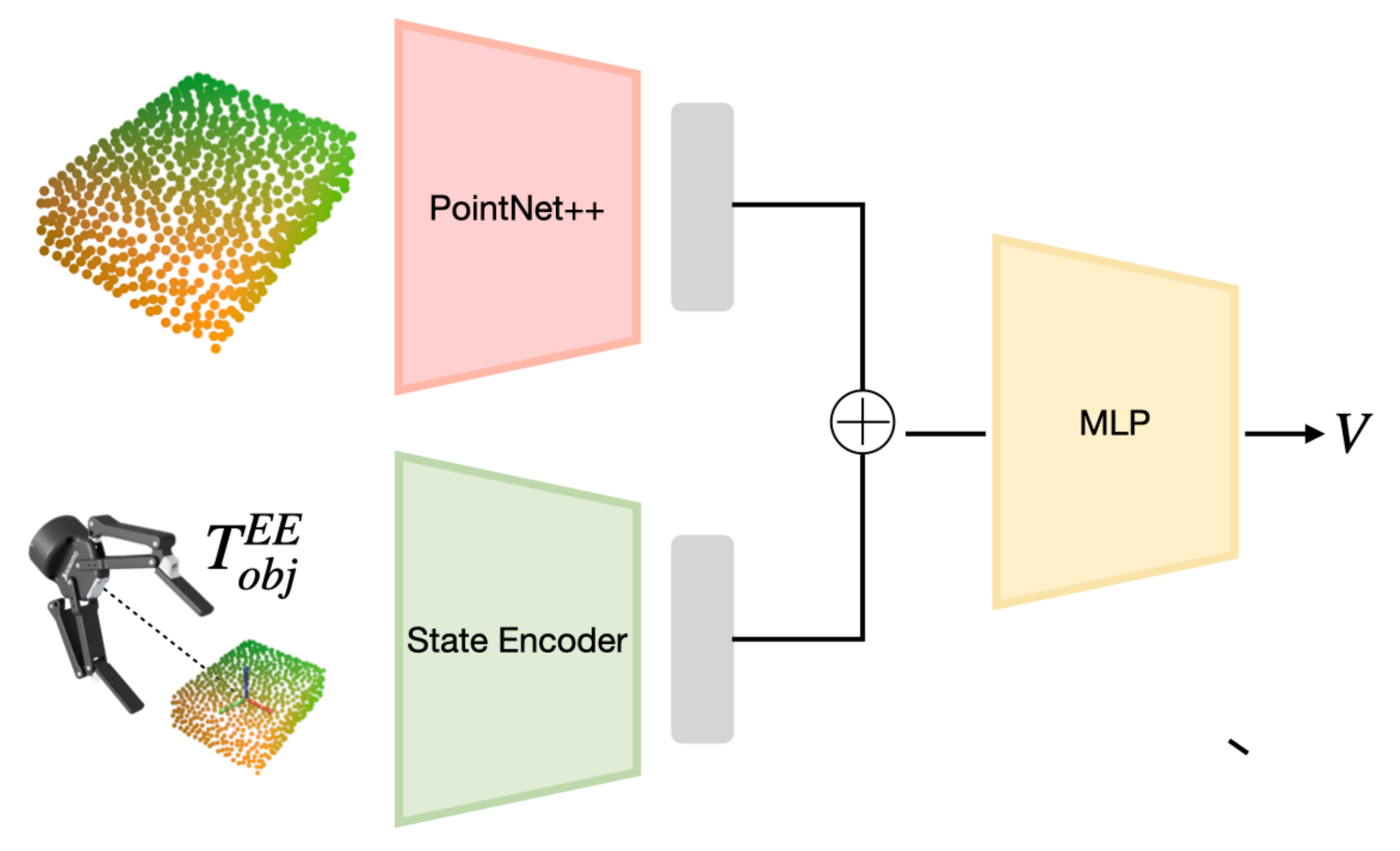}
    \caption{\ourmethod~trains a value function to serve as a cost function within MPC. The PointNet++ encoder takes a segmented point cloud as input, and the state encoder takes an end-effector pose with respect to the mean of the object point cloud $T^{EE}_{obj}$ as input. Features from each encoder are concatenated and then used as input to the MLP head to estimate the value.}
    \label{fig:model_arch}
\end{figure}

The value function architecture consists of three main components: the PointNet++ encoder, the state encoder, and the value head network (see Figure~\ref{fig:model_arch}).
The PointNet++ encoder employs three sets of abstraction layers, followed by three fully connected layers.
The first set abstraction layer uses furthest point sampling (FPS) to reduce the input to $256$ points.
It then applies a grouping query within a radius of $3cm$, selecting up to $64$ points per group.
This is followed by a local PointNet consisting of three fully connected layers of size $3$, $64$, and $128$.
The second set abstraction layer further downsamples the point cloud to $64$ points, using a grouping query that captures up to $128$ points within a $4cm$ radius.
Its local PointNet has fully connected layers of sizes $128$, $128$, and $256$.
The third abstraction layer skips FPS, instead grouping all points together, and employs a local PointNet with layer sizes of $256$, $256$, and $512$.
After the set abstraction layers, the output passes through three fully connected layers with sizes $512$, $256$, and $128$. 
Between these layers, layer normalization, dropout with $p=0.2$, and ELU activations are applied.
The final output of the PointNet++ encoder is a $128$ dimensional feature vector.

The state encoder processes a $9$-dimensional input that encodes the pose of the robot's end-effector. 
The first three dimensions represent the translational components of the pose, while the remaining six dimensions correspond to the first two columns of the end-effector's rotation matrix (6D Gram-Schmidt) for its orientation~\cite{pravdova2024representation}.
The state encoder consists of two hidden layers with sizes $64$ and $32$, producing a $32$-dimensional embedding vector.
This output is concatenated with the $128$-dimensional feature vector from the PointNet++ encoder, resulting in a $160$-dimensional feature vector.
Finally, the value head network consists of fully connected layers with three hidden layers of sizes $256$, $128$, and $64$.

\subsection{MPC details}
\label{appendix:mpc}
We use a sampling-based MPPI optimizer with three main constraints: (1) joint limits on position, velocity, acceleration, and jerk; (2) self-collision avoidance; and (3) robot-world collision avoidance. 
The cost function used in \ourmethod~is a grasp cost using the learned value function and the default CuRobo cost terms:

\begin{equation} C_{\text{\ourmethod}} = C_{\text{curobo}} + \omega C_{\text{grasp}} \end{equation}
where $\omega=1000$ and $C_{\text{curobo}}$ consist of a world collision cost, a self-collision cost, and a bounds cost for trajectory smoothness and keeping the joint states within limits.

To encourage smooth control sequences, we sample actions via a Halton sequence and fit degree-5 B-splines.
Hyperparameters of MPC used in our experiments are described in Table~\ref{table:mpc_parameters}.

\begin{table}[h]
\centering
\caption{CuRobo MPC Hyperparameters}
\begin{tabular}{l r} 
 \toprule
 \textbf{Parameter} & \textbf{Value}  \\ 
 \toprule
 horizon & $30$ \\
 control\_space & ACCELERATION \\
 init\_cov & $0.5$ \\
 gamma & $0.98$ \\
 \midrule
 n\_iters & $2$ \\
 cold\_start\_n\_iters & $5$ \\
 step\_size\_mean & $0.9$ \\
 step\_size\_cov & $0.5$ \\
 beta & $1.0$ \\
 alpha & $1.0$ \\
 \midrule
 num\_particles & $400$ \\
 update\_cov & True \\
 kappa& $0.01$ \\
 null\_act\_frac& $0.05$ \\
 \midrule
 sample\_mode & BEST \\
 best\_action & REPEAT \\
 squash\_fn& CLAMP \\
 n\_problems & $1$ \\
 random\_mean & True \\
 use\_coo\_sparse & True \\
 \midrule
 sample\_ratio:  halton & $0.3$ \\
 sample\_ratio: halton-knot & $0.7$ \\
 sample\_ratio: random & $0$ \\
 sample\_ratio: random-knot & $0$ \\
 \bottomrule
\end{tabular}
\label{table:mpc_parameters}
\end{table}

\subsection{Simulated Environment Setup}

\begin{figure*}
    \centering
    \includegraphics[width=\linewidth]{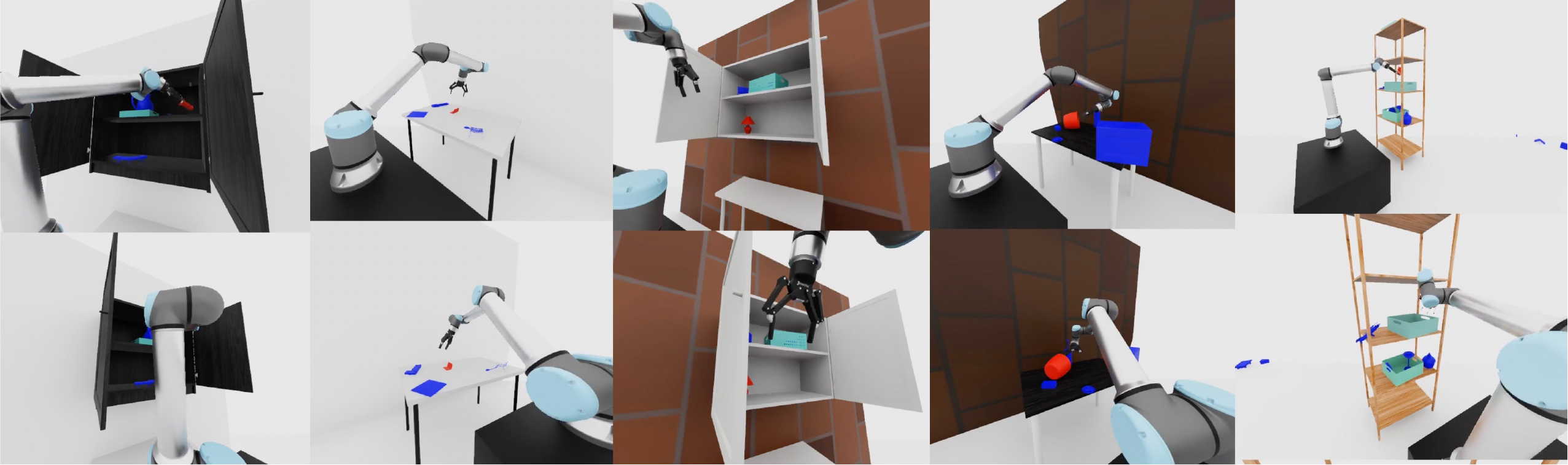}
    \caption{Representative scenes in FetchBench~\cite{han2024fetchbench} for grasping in clutter. We replace Isaac Gym~\cite{makoviychuk2021isaac} used as the underlying simulator in the original FetchBench with Isaac Sim to simulate a Robotiq 2F-140 gripper.}
    \label{fig:fetchbench_additional}
\end{figure*}
In this work, we evaluate \ourmethod~and the competitive baselines in FetchBench~\cite{han2024fetchbench}.
However, the original FetchBench environment is incompatible with the Robotiq 2F-140 gripper due to its closed-loop kinematic chain, and its simulation platform, Isaac Gym~\cite{makoviychuk2021isaac}, has been deprecated. 
To address these limitations, we replace Isaac Gym with Isaac Sim, enhancing simulation accuracy, ensuring compatibility with the Robotiq 2F-140 gripper, and benefiting from improved support and ongoing development (see Figure~\ref{fig:fetchbench_additional}).

In the experiments, we assess \ourmethod~and the baseline methods over $90$ unique scenes, each containing $60$ problems, resulting in a total of $5,400$ test problems. 
These include $1,415$ on-table, $2,424$ on-shelf, $744$ in-basket, and $817$ in-drawer cases.

\subsection{Real-world Environment Setup}
\begin{figure}
    \centering
    \includegraphics[width=0.98\linewidth]{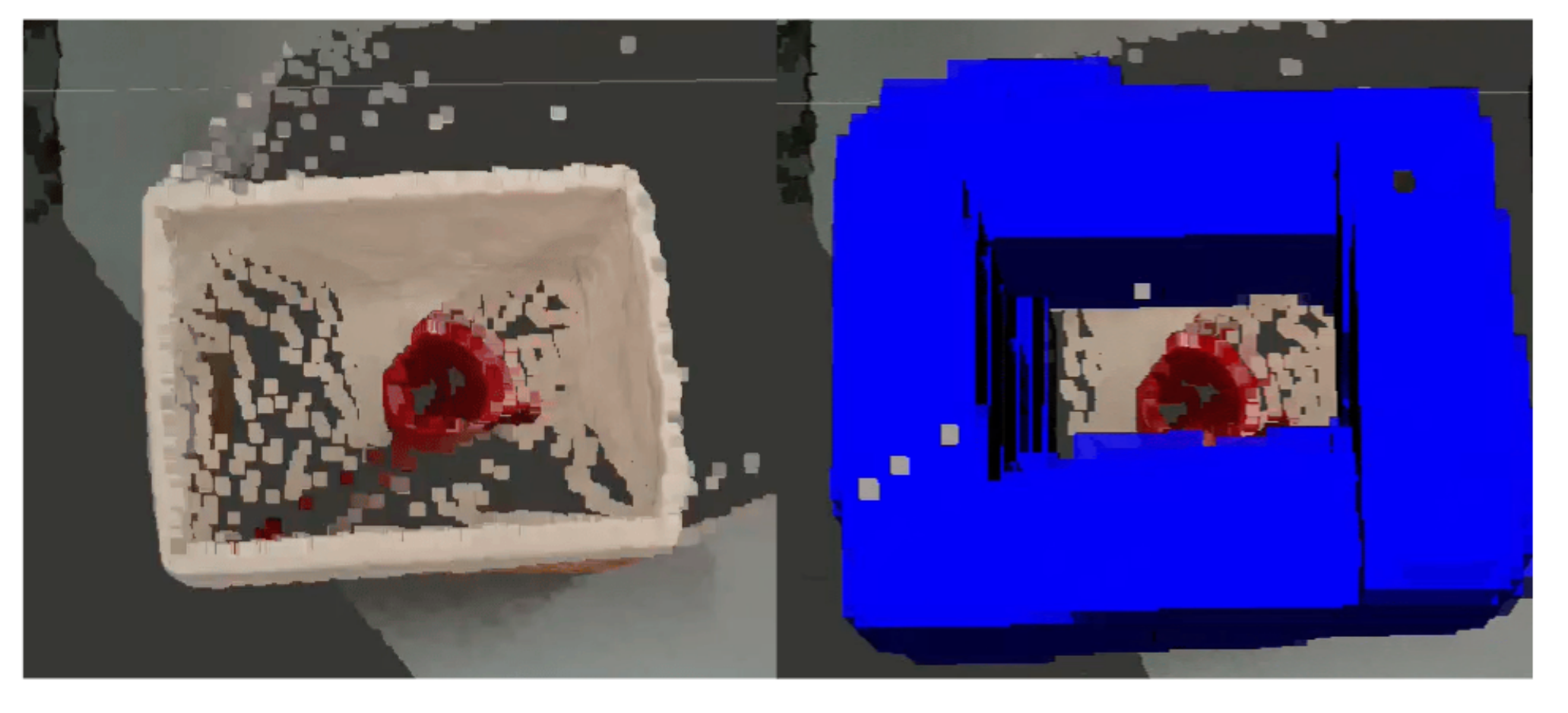}
    \caption{Example collision voxel visualization using NVBlox. By leveraging NVBlox, a GPU-accelerated SDF library, \ourmethod~successfully avoids collision and grasp the target object.}
    \label{fig:nvblox}
\end{figure}

The real-world experiments use SAM-Track~\cite{kirillov2023segment}, which combines Grounding-DINO~\cite{zhang2022dinodetrimproveddenoising} for object detection and SAM~\cite{kirillov2023segment} for segmentation, to track the target object.
The segmented depth image is then projected to a point cloud, and latent observation features are obtained from our observation encoder. This latent feature is then sent to our MPC, which is implemented as a ROS Python node.

We use ROS control to switch between a trajectory controller and a velocity controller. 
We use cuRobo's motion planner to move to the approach pose and then call our MPC to take over and grasp the object. 
To avoid collision, we use Nvblox~\cite{millane2024nvbloxgpuacceleratedincrementalsigned} to represent the scene for both motion planning and MPC with the resolution of $1\text{cm}$ voxels. (see Figure~\ref{fig:nvblox})
We run MPC for $120$ steps and then close the gripper. 
Once the gripper is closed, we then lift the gripper $10\text{cm}$ along the opposite direction of the gravity vector using the same motion planner.

\begin{figure}
    \centering
    \includegraphics[width=0.98\linewidth, trim={0 5cm 0 0}, clip]{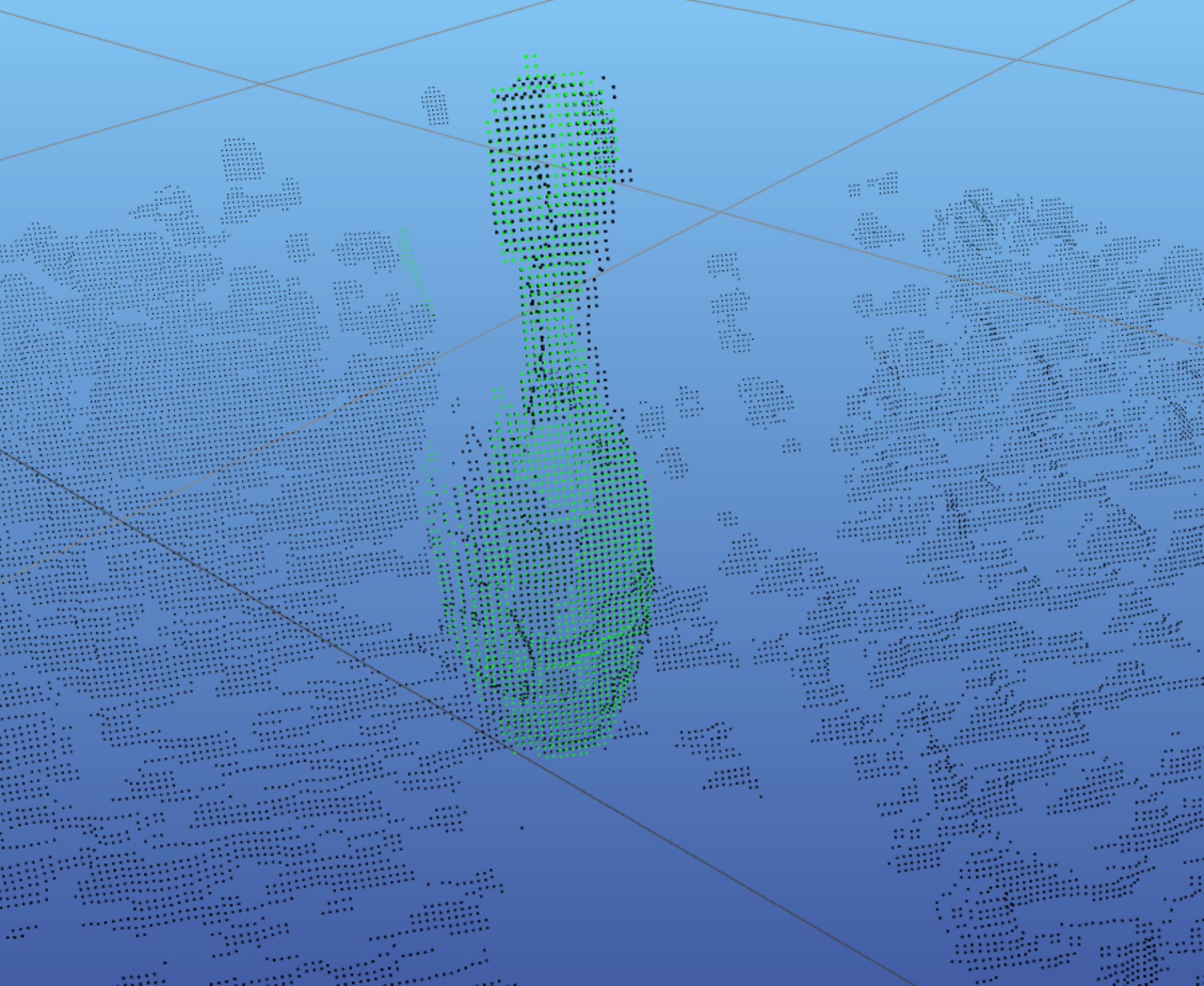}
    \caption{Overlaying object point cloud across different methods. To maintain identical experimental conditions, the current point clouds (\textcolor{green}{green}) are overlaid on the reference object point clouds (black).}
    \label{fig:overlay}
\end{figure}

\textbf{Identical conditions ensure fair comparison between methods.} 
To ensure fair comparison, we maintain identical experimental conditions across both approaches: the target grasp pose, pre-grasp pose, and robot configuration at the pre-grasp position remain the same between methods. 
Object pose consistency is verified by overlaying the point clouds of the target object across different method evaluations, ensuring that any performance differences stem from the grasping approach rather than variations in object positioning (see Figure~\ref{fig:overlay}).

\subsection{Implementation Details of Baseline Methods}
\textbf{GraspAPI.}
We utilize the grasp planning API provided by CuRobo~\cite{sundaralingam2023curobo} to generate a trajectory for grasping. 
Initially, the API plans trajectories from the robot's starting position to a set of candidate grasp poses, selecting the shortest path among these options. 
Once the desired grasp pose is identified, a pre-grasp pose is computed by applying a fixed linear offset to the selected grasp pose.
Then, the API generates a trajectory from the robot's initial position to the pre-grasp pose. 
Following this, the API employs constrained motion planning to produce a linear trajectory between the pre-grasp pose and the grasp pose. 
Lastly, the API returns two trajectories: one from the robot's initial position to the pre-grasp pose, and another from the pre-grasp pose to the desired grasp pose.

\textbf{Operational Space Control.}
Similar to the baseline approach used in FetchBench~\cite{han2024fetchbench}, we use operational space control to move the end effector linearly from the pre-grasp to the grasp pose.

\textbf{Diffusion Policy.}
\label{appendix:diffusion}
We train a diffusion policy~\cite{chi2023diffusion} using behavior cloning (BC)
The diffusion policy uses a conditional Denoising Diffusion Probabilistic Models (DDPM), and its conditional noise prediction model employs the CNN-based architecture introduced in \cite{chi2023diffusion}.
This architecture is a $3$-level UNet architecture consisting of conditional residual blocks with channel dimensions $[128, 256, 512]$. 
The diffusion time step is encoded by a multi-layer perceptron (MLP) to produce a $128$-dimensional embedding, which is concatenated with the visual embedding from the PointNet++ encoder and the state embedding from the state encoder.
This concatenation yields a $288$-dimensional conditional vector.
The architecture of the PointNet++ encoder and the state encoder shares the same architecture as that used in the value function network in \ourmethod.
In the experiments, we use an action horizon of $4$ and an observation history length of $2$.
During training, we use $100$ diffusion steps.
For faster inference during evaluation, we employ Denoising Diffusion Implicit Models (DDIM) with $5$ diffusion steps.

The synthetic dataset contains both successful and failed trajectories, corresponding to feasible and infeasible grasp poses. 
To train the diffusion policy, we utilize only the successful trajectories from the dataset.
The policy produces $6$-dimensional action outputs, where the first three dimensions represent the change in position, and the last three correspond to the change in the end-effector's axis-angle representation.
Expert actions in the dataset are derived by computing the difference between the current and next end-effector poses. Since the motion planner generates a fine-grained trajectory during data collection, we aggregate actions by skipping $4$ frames and computing the combined pose difference over this interval.

\textbf{Transformer Policy.}
We train a transformer-based policy using BC from only successful trajectories similar to the diffusion policy described in Appendix~\ref{appendix:diffusion}. 
To train the transformer policy, we utilize only the successful trajectories from the dataset.
Our transformer policy, similar to the one utilized in FetchBench, is built upon the OPTIMUS architecture.
Consistent with the value function in \ourmethod, the transformer policy processes a segmented target object point cloud alongside the robot end-effector pose relative to the center of the target object point cloud.
To encode visual input, we utilize a PointNet++ encoder with the same architecture as the one used in our value function.
The transformer policy uses the same action space and expert actions for training as the Diffusion Policy.

\textbf{Implicit Q-Learning.}
We train a policy using Implicit Q-Learning, the state-of-the-art offline RL approach.
In this training, the objective is to maximise the cumulative discounted rewards.
Thus, instead of using a cost function (Eq.~\ref{eq:cost}) defined for \ourmethod, we use the following reward function:

\begin{equation}
    r_{t} = 
    \begin{cases}
        1 & |q_{goal, i} - q_{t,i}| \leq 5e^{-3}, \ \forall i, \ and \ \mathbbm{1}_{\text{feasible}}=1, \\
        0 & \text{Otherwise}
    \end{cases}
\end{equation}
In IQL, both the state-based value function and the Q-value function are trained.
We employ the same network architecture for these value functions as that used in the value function of \ourmethod, with the key difference that the Q-value function incorporates an action as part of its input, which is included in the input to the state encoder.
The policy architecture is similar to the value function architecture; however, it is parameterized as a Gaussian distribution with a fixed, state-independent standard deviation.
The policy outputs the mean of the Gaussian distribution for an action space with $6$ dimensions.

\end{document}